\documentclass[10pt,twocolumn,letterpaper]{article}

\usepackage[pagenumbers]{cvpr} % To force page numbers, e.g. for an arXiv version

\usepackage[dvipsnames]{xcolor}

\usepackage{multirow}
\newcommand{\tableCellHeight}{1}
\newcommand{\tabstyle}[1]{
  \setlength{\tabcolsep}{#1}
  \renewcommand{\arraystretch}{\tableCellHeight}
  \centering
  \small
}
\usepackage[misc]{ifsym}
\usepackage[hang]{footmisc}
\newcommand\blfootnote[1]{%
\begingroup
\renewcommand\thefootnote{}\footnote{#1}%
\addtocounter{footnote}{-1}%
\endgroup
}

\definecolor{cvprblue}{rgb}{0.21,0.49,0.74}
\usepackage[pagebackref,breaklinks,colorlinks,citecolor=cvprblue]{hyperref}
\usepackage{xcolor}         % colors
\definecolor{myred}{RGB}{255, 94, 111}
\definecolor{myyellow}{RGB}{255,196,0}%{183, 146, 48}
\definecolor{mygreen}{RGB}{62,141,39}
\usepackage{colortbl}

\def\paperID{5685} % *** Enter the Paper ID here
\def\confName{CVPR}
\def\confYear{2024}

\newcommand{\name}{DualDETR}
\title{Dual DETRs for Multi-Label Temporal Action Detection}

\author{
Yuhan Zhu\textsuperscript{1}\textsuperscript{*} \quad
Guozhen Zhang\textsuperscript{1}\textsuperscript{*} \quad
Jing Tan\textsuperscript{2} \quad
Gangshan Wu\textsuperscript{1} \quad
Limin Wang\textsuperscript{1,3,\dag}\\
\textsuperscript{1}State Key Laboratory for Novel Software Technology, Nanjing University \\
\textsuperscript{2}The Chinese University of Hong Kong \quad
\textsuperscript{3}Shanghai AI Lab
\\
\textbf{\normalsize\url{https://github.com/MCG-NJU/DualDETR}}
}

\begin{document}
\maketitle

\begin{abstract}
Temporal Action Detection (TAD) aims to identify the action boundaries and the corresponding category within untrimmed videos. Inspired by the success of DETR in object detection, several methods have adapted the query-based framework to the TAD task. However, these approaches primarily followed DETR to predict actions at the instance level (i.e., identify each action by its center point), leading to sub-optimal boundary localization. To address this issue, we propose a new Dual-level query-based TAD framework, namely~\name, to detect actions from both instance-level and boundary-level.
Decoding at different levels requires semantics of different granularity, therefore we introduce a two-branch decoding structure. This structure builds distinctive decoding processes for different levels, facilitating explicit capture of temporal cues and semantics at each level.
On top of the two-branch design, we present a joint query initialization strategy to align queries from both levels. Specifically, we leverage encoder proposals to match queries from each level in a one-to-one manner. Then, the matched queries are initialized using position and content prior from the matched action proposal. 
The aligned dual-level queries can refine the matched proposal with complementary cues during subsequent decoding.
We evaluate~\name~on three challenging multi-label TAD benchmarks. The experimental results demonstrate the superior performance of~\name~to the existing state-of-the-art methods, achieving a substantial improvement under det-mAP and delivering impressive results under seg-mAP.
\end{abstract}
\blfootnote{*: Equal Contribution.~~~\dag: Corresponding author (lmwang@nju.edu.cn).}

%------------------------------------------------------------------------------%

\section{Introduction}
Temporal Action Detection (TAD)~\cite{ssn, BSN, bmn, TAL_Net, rapnet, afsd, rtdnet,g_tad} is one of the fundamental tasks in video understanding~\cite{simonyan2014twostream,yuan2017temporal,tsn,zhang2023extracting,yang2019step,zhao2023asymmetric,ding2022motion, ding2022dual, ding2023prune, li2020actions}, with a wide range of real-world applications in video editing~\cite{huang2020movienet}, sports analytics~\cite{Li_2021_ICCV, giancola2018soccernet}, surveillance footage analysis~\cite{xu2019localization}, and autonomous driving~\cite{9857383}. TAD aims to identify the starting and ending time of human actions, and simultaneously recognize the corresponding action categories. 
To address the complex real-world application scenario for TAD, we focus on the complicated Multi-label Temporal Action Detection (Multi-label TAD)~\cite{pointtad, mlad, ctrn, ms_tct,pdan}, where diverse actions from different categories co-exist in untrimmed videos, often with {\em significant temporal overlaps}.

\begin{figure}[t]
\centering
\includegraphics[width=1\linewidth]{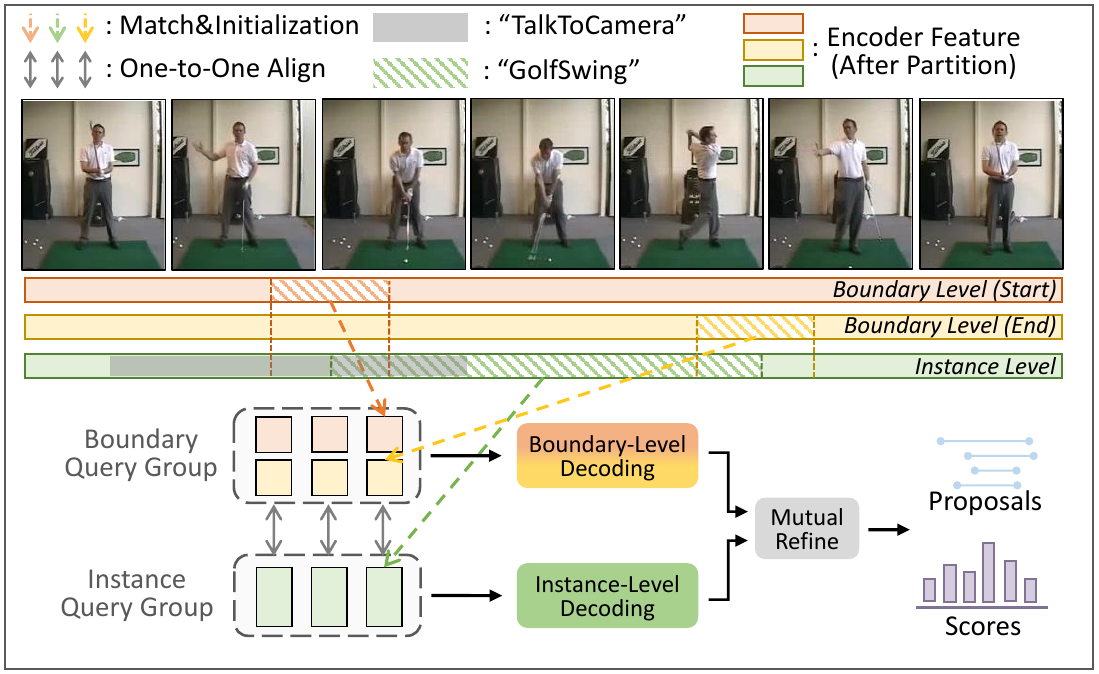}
\caption{\textbf{\name} operates at both the instance level and boundary level (start, end) by using two groups of queries, with each group corresponding to one level. To capture specific semantics at each level, we introduce a two-branch decoding structure. This structure separates the decoding process for each level, allowing queries from each group to focus on their corresponding encoder feature map. Furthermore, we propose a query alignment strategy equipped with joint initialization. This strategy aligns the queries from two groups by matching them with the same detection goal, as denoted by the bidirectional arrow.}
\vspace{-2mm}
\label{fig:teaser}
\end{figure}

Inspired by the success of DETR~\cite{DETR} in object detection, several approaches~\cite{rtdnet, tadtr, react, temporal_perceiver, pointtad, self_detr} adopted the query-based detection pipeline and used a sparse set of learnable decoder queries to directly predict actions without NMS post-processing. These approaches commonly follow DETR to detect actions from the instance level. They identify each action by its center point and predict duration based on offsets.
Although instance-level detection benefits the capture of the important semantic frames within actions, such practice overlooks a crucial gap between two tasks: object locations are mainly decided by their centroids, whereas actions in videos are often defined by the starting and ending boundaries. Hence, these methods remain poor at the precise localization of action boundaries.

To bridge this gap, we propose a novel Dual-level query-based TAD framework (\name) that integrates both instance-level and boundary-level modeling into the action decoding. As depicted in~\cref{fig:teaser},~\name~employs two groups of decoder queries, namely boundary-level query group (\textcolor{myred}{red} and \textcolor{myyellow}{yellow}) and instance-level query group (\textcolor{mygreen}{green}), with each group corresponding to one level of decoding. The instance-level queries capture important semantic frames within the proposal, providing a holistic understanding of the action content. Meanwhile, the boundary-level queries focus on the details around proposal boundaries, exhibiting higher sensitivity to the salient boundary frames. 
Following the dual-level decoding pipeline,~\name~can improve the action proposals by combining reliable recognition from the instance level and precise boundary refinement from the boundary level.

Simply decoding the two levels of queries via a shared decoder does not yield optimal performance. In general, decoding from boundary and instance levels requires semantics of different granularity. Using a shared decoder for dual levels will fail to focus on specific semantics at each level, hence hampering the effective decoding for both levels. 
To address this, we propose the {two-branch decoding structure with feature partition} to use a distinctive decoder for each level. 
Specifically, we partition the encoder feature map along the channel dimension to represent the boundary (start, end) and instance levels. 
The separation benefits the explicit capture of individual characteristics at each level. This design is especially helpful in multi-label TAD scenarios where different action instances overlap. 
For instance, as depicted in ~\cref{fig:teaser}, the action ``{\em GolfSwing}'' starts while the action ``{\em TalkToCamera}'' is ongoing in the background. Under such complex scenario, it is challenging to accurately determine the boundaries of each action.
Feature separation enables explicit cues for each action at each level to be preserved and processed in different feature maps, thus benefiting the precise localization of overlapping actions. 

With the two-branch design for dual levels, we present a novel joint query initialization strategy to align queries from both levels and achieve complementary refinement of action proposals during subsequent decoding.
First, we establish the alignment from action proposals predicted by the encoder. Each action proposal is paired with a starting boundary query, an ending boundary query, and an instance query. This alignment allows for a one-to-one matching between boundary and instance queries, enabling joint updates of the matched proposal during decoding.
Second, similar to \cite{dab_detr,dino}, each query is constructed as a pair of position and content vectors. On top of this, instead of learning sample-agnostic priors from training~\cite{rtdnet,tadtr}, the position and content vectors are initialized with the position and semantic priors from their matched proposal. 
Thanks to the joint query initialization, the position vectors guide the queries to explicitly focus on the matched proposal, while the content vectors provide semantic guidance for both pair-wise relation modeling and global feature refining.

We conduct extensive experiments on three challenging multi-label TAD benchmarks, MultiTHUMOS~\cite{multi_thumos}, Charades~\cite{charades} and TSU~\cite{dai2022toyota}. Our proposed~\name~outperforms the previous state-of-the-art methods by a large margin under detection-mAP, demonstrating its fine-grained recognition and precise localization abilities. Notably,~\name~showcases impressive per-frame detection accuracy under the segmentation-mAP, comparing with both detection-based methods and segmentation-based methods.

In summary, our contributions are threefold:
\begin{itemize}
    \item We identify the sub-optimal localization issue from the instance-level detection paradigm in previous query-based TAD approaches and present a novel dual-level query-based action detection framework (\name).
    \item To facilitate effective dual-level decoding, we devise a two-branch decoding structure and joint query initialization strategy to align dual-level queries and refine proposals with complementary efforts.
    \item Extensive experiments demonstrate that~\name~surpasses the previous state-of-the-art on three challenging benchmarks under det-mAP and achieves impressive results under seg-mAP, compared with both detection methods and segmentation methods.
\end{itemize}

%------------------------------------------------------------------------------%
\section{Related Work}
\noindent\textbf{Multi-Label Temporal Action Detection.} Prior studies~\cite{sardari2023pat} in multi-label TAD have primarily formulated the problem as a frame-wise classification (segmentation) task, with an emphasis on action class recognition rather than precise action boundary localization for all action instances. Early research~\cite{piergiovanni2018superevent,piergiovanni2019TGM} sought to capture temporal context through carefully designed Gaussian kernels. Other works captured and modeled temporal relations with dilated attention layers~\cite{pdan} or a combination of convolution and self-attention blocks~\cite{ms_tct}. Coarse-Fine~\cite{coarse_fine} adopted a two-stream architecture, facilitating the extraction of features from distinct temporal resolutions. MLAD~\cite{mlad} leveraged the attention mechanism to model actions occurring at the same and across different time steps. PointTAD~\cite{pointtad} marked a return of multi-label TAD to the domain of action detection task~\cite{pointtad, shao2023action, shi2023temporal, gao2023haan}. In this paper, we present a dual-level framework to further explore the potential of the query-based framework, with a specific focus on the precise localization of action instances in the multi-label TAD task.

\noindent\textbf{Boundary Information in TAD.} 
Previous studies~\cite{shi2023tridet, xia2022learning, nag2022proposal, wang2022rcl} on action boundaries primarily focused on extracting high-quality boundary features for proposal generation or evaluation. The early methods~\cite{ssn,tsi, BSN,bmn,dbg} employed convolutional networks to extract boundary features. MGG~\cite{mgg} refined proposal boundaries by identifying positions with higher boundary scores. Temporal ROI Align~\cite{mask_rcnn} or boundary pooling techniques were adopted by TCANet~\cite{tcanet} and AFSD~\cite{afsd} to retrieve features for boundary refinement.
Regarding the query-based methods, RTDNet~\cite{rtdnet} multiplied the boundary scores with the original video features. However, RTDNet encountered difficulties in achieving reliable recognition scores thus leading to unsatisfactory detection performance, leaving the appropriate way to incorporate boundary information into the query-based framework as an open problem.
In this paper, we aim to tackle this problem by proposing a dual-level framework to carefully addresses these challenges with its design. 

\noindent\textbf{Query Formulation in DETR.} The formulation of decoder queries was widely studied in the objection detection domain. DETR~\cite{DETR} utilized randomly initialized object queries during training to learn dataset-level object distribution. Anchor DETR~\cite{wang2109anchor} initialized queries based on anchor points to establish a specific detection mode. Deformable DETR~\cite{conditional_detr_v2} and Conditional DETR v2~\cite{conditional_detr_v2} leveraged the encoder proposals to provide positional priors for decoder queries. DAB-DETR~\cite{dab_detr} formulated decoder queries with a content vector and an action vector. Upon this, DINO~\cite{dino} incorporated position priors for the position vector and randomly initialized the content query during training. In this paper, we share a distinct motivation with the aforementioned object detection methods, which is to achieve effective alignment between dual-level queries.

%------------------------------------------------------------------------------%

\begin{figure*}[t]
\centering
\includegraphics[width=1\linewidth]{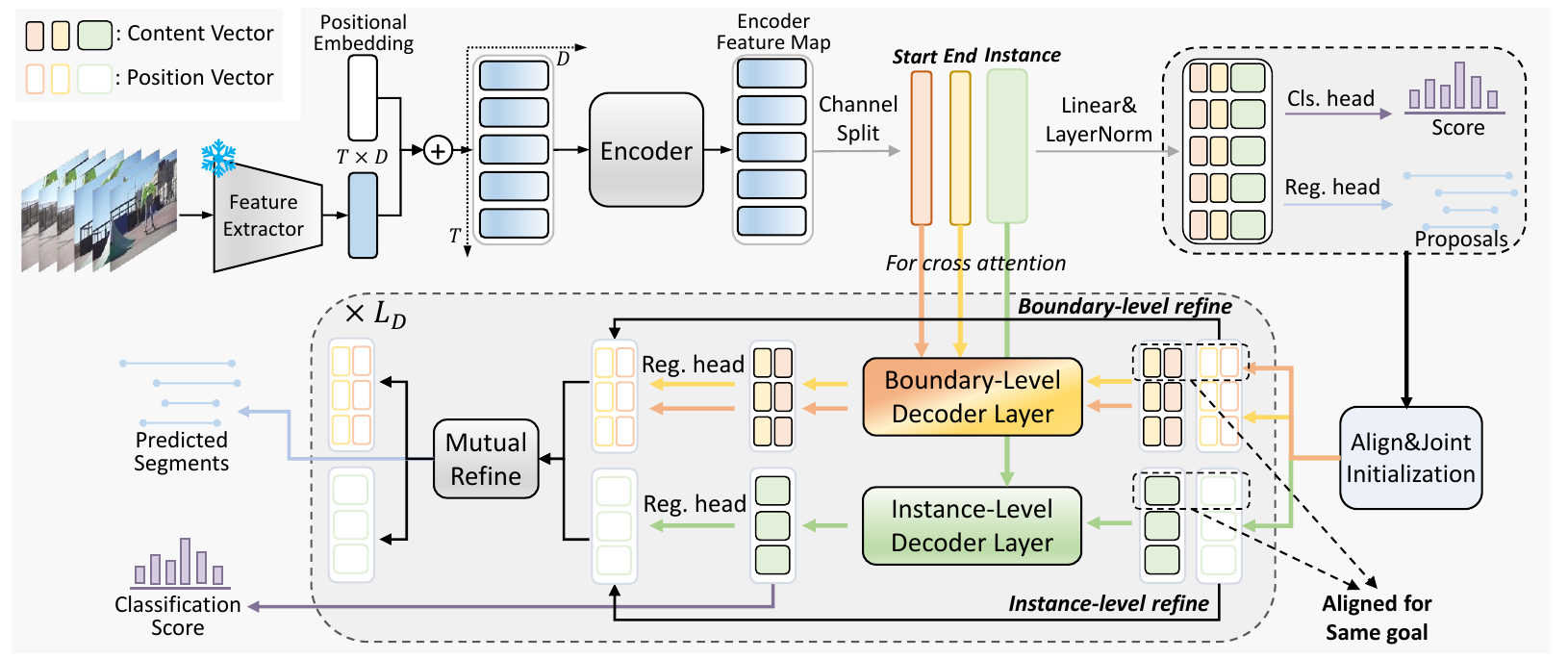}
\caption{\textbf{Pipeline of~\name.} The pre-extracted video features, augmented with the positional embedding, pass through a transformer encoder to produce the encoder feature map. This map is divided along the channel dimension into separate feature maps for the boundary-level (start, end) and instance-level modeling, respectively.
An auxiliary dense detection head is applied to generate encoder proposals and scores. Upon this, decoder queries are constructed using the query alignment strategy. The decoding process is performed at dual levels. Thanks to the query alignment, dual-level queries can perform a complementary refinement through the mutual refinement module. Finally,~\name~directly output action instance predictions without NMS post-processing.}
\label{fig:overview}
\end{figure*}

\section{Method}
\subsection{Preliminaries}
\textbf{Query-Based TAD Framework}~\cite{rtdnet, tadtr, react, temporal_perceiver, pointtad, self_detr} is proposed inspired by the success of DETR~\cite{DETR}. It employs a transformer architecture~\cite{transformer} and typically consists of an encoder and a decoder. The encoder takes video features $X\in \mathbb{R}^{T \times D}$ as input, which are extracted by a pre-trained video encoder (e.g., I3D~\cite{i3d}), where $T$ and $D$ represent the temporal length and feature dimension, respectively. The encoder employs self-attention to model snippet-level temporal relation. Following the refinement by $L_E$ encoder layers, the decoder employs $N_q$ action queries to simultaneously model action-level relations using self-attention and refine global features using cross-attention. Subsequently, a detection head is applied to these action queries to obtain sparse detection results without post-processing technique like Non-Maximum Suppression (NMS). During training, optimal bipartite matching is performed between predicted and ground truth action instances, enabling the calculation of classification and localization losses.

\noindent\textbf{Deformable Attention}~\cite{deformable_detr} is proposed to address the slow convergence issue of DETR while improving its computational efficiency. In this paper, we incorporate deformable attention as a tool to explicitly guide attention localization.
Let $q$ index a query element. Given the query feature $\boldsymbol{z}_q$, a 1-d reference point $\boldsymbol{t}_q \in [0,1]$, and the input feature map $X\in \mathbb{R}^{T \times D}$, the deformable attention is calculated as:
\begin{equation}
\begin{aligned}
% \begin{small}
\text{DeformAttn}&(\boldsymbol{z}_q, \boldsymbol{t}_q, X)=\\
\sum_{m=1}^M \boldsymbol{W}_m & \left[\sum_{k=1}^K A_{m q k} \cdot \boldsymbol{W}_m^{\prime} X\left(\boldsymbol{t}_q+\Delta \boldsymbol{t}_{m q k}\right)\right].
% \end{small}
\label{eq:deform-attn}
\end{aligned}
\end{equation}
Here, $A_{m q k}$ represents the attention weight computed as $\text{SoftMax}(\text{Linear}(\boldsymbol{z}_q))$. $m$ indexes the attention head and $k$ indexes the sampling temporal points. $M$ and $K$ denote the number of attention heads and sampling points, respectively. $\Delta \boldsymbol{t}_{m q k}$ represents a normalized 1-d sampling offset.

\subsection{Overview}
Given an untrimmed video,~\name~aims to predict a set of action instances $\Psi=\left\{\varphi_n=\left(t^{s}_{n}, t^{e}_{n}, a_n\right)\right\}_{n=1}^{N_g}$. Here, $N_g$ represents the number of ground-truth action instances, $t^{s}_{n}, t^{e}_{n}$, and $a_n$ denote the starting, ending time and the corresponding action label of action instances.

The entire pipeline is illustrated in~\cref{fig:overview}. \name~operates on video features $X\in \mathbb{R}^{T \times D}$, that are extracted by a pre-trained feature extractor (e.g., I3D~\cite{i3d}). The model employs the encoder-decoder pipeline. For feature encoding, the model uses a transformer encoder with deformable attention to efficiently perform temporal modeling at the snippet level. 
For action decoding, we introduce a {\bf two-branch decoding structure} based on transformer decoders to predict actions from both boundary and instance levels. Accordingly, decoder queries are divided into two groups and the encoder features are also divided along channels for dual-level cross-attention. At each branch, the decoder takes in the corresponding queries and features to make predictions. To achieve complementary refinement of proposals from both levels, we propose a {\bf joint query initialization strategy }to align different groups of queries based on the action proposals predicted from the encoder. Each proposal is matched with a pair of boundary queries and one instance query. The content and position vectors of the queries are initialized by the feature embedding and boundary position of the matched proposal in correspondence. 
At the end of each layer, a {\bf mutual refinement module} facilitates the communication between the aligned queries. Finally, the classification scores generated by the instance-level content vector, along with the proposals from the mutual refinement module, serve as the final detection results, without the need for NMS post-processing.

\subsection{Dual-Level Query Construction}
In this subsection, we present the construction of the decoder queries in our dual-level framework. We divide the decoder queries into two groups, one group for the boundary decoding branch and the other for the instance branch. Similar to~\cite{dab_detr, dino, adamixer}, we disentangle the position and content decoding for each query by constructing it as a pair of position and content vectors. 
The instance-level query group, denoted as $\boldsymbol{i}$, consists of content matrix $\boldsymbol{i}^{con} \in \mathbb{R}^{N_q \times D/2}$ and  position matrix $\boldsymbol{i}^{pos} \in \mathbb{R}^{N_q \times 2}$, where $N_q$ is the number of queries, and $D$ denotes the number of feature channels. The content vector captures high-level semantic information, while the position vector contains two normalized scalars representing the center and duration of a proposal. Similarly, the boundary-level query group consists of start and end queries, represented as $\boldsymbol{s}$ and $\boldsymbol{e}$ respectively. Each boundary query also contains content and position matrix, denoted as $\boldsymbol{s}^{con}, \boldsymbol{e}^{con} \in \mathbb{R}^{N_q \times D/4}$ and $\boldsymbol{s}^{pos},\boldsymbol{e}^{pos} \in \mathbb{R}^{N_q \times 1}$. The position vectors contain normalized scalars representing the starting and ending times of a proposal. During the decoding process, the position vectors serve as reference points, providing explicit positional guidance in both self-attention and cross-attention. Meanwhile, the content vectors offer semantic guidance for pair-wise query relation modeling in self-attention and query refinement in cross-attention. This dual-level query corresponds to the subsequent two-barnch decoding.

\subsection{Query Alignment with Joint Initialization}
\begin{figure}[t]
\centering
\includegraphics[width=1\linewidth]{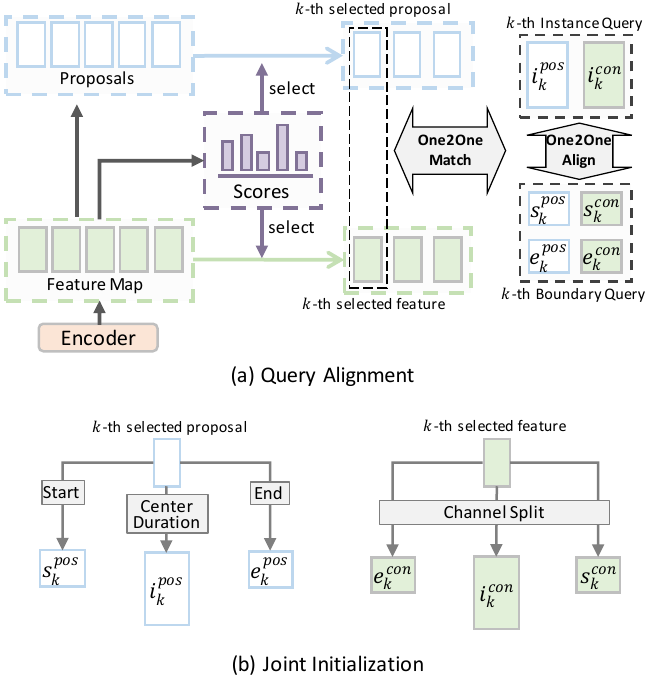}
\caption{\textbf{Query Alignment with Joint Initialization.} (a) Instance queries and boundary queries are aligned to match with the encoder predictions in a one-to-one manner. (b) The matched encoder prediction serves as the initialization for dual-level queries.}
\vspace{-4mm}
\label{fig:query_init}
\end{figure}

After constructing the action queries with two groups and decoding each group within separate branches, it is important to align the queries from both groups to facilitate their joint refinement of action proposals. This alignment enables the model to benefit from both the instance-level queries, which provide semantic guidance for recognition, and the boundary-level queries, which refine the proposal boundaries with high precision. To achieve query alignment, we first obtain proposals and classification scores by applying a detection head to the encoder feature map. These proposals, selected based on their classification scores, are then matched with the decoder queries from both groups. For example, considering the $k$-th selected proposal, as depicted in~\cref{fig:query_init}~(a), we match this proposal with the $k$-th instance-level query $\boldsymbol{i}_k=\{\boldsymbol{i}^{pos}_k,\boldsymbol{i}^{con}_k \}$, as well as the $k$-th boundary-level query $\boldsymbol{s}_k=\{\boldsymbol{s}^{pos}_k,\boldsymbol{s}^{con}_k \}, \boldsymbol{e}_k=\{\boldsymbol{e}^{pos}_k,\boldsymbol{e}^{con}_k \}$. This matching process ensures an one-to-one alignment between the instance and boundary queries, allowing them to jointly update the matched proposal during the decoding process.

Based on the matching, we propose a joint query initialization strategy to provide a good kick-start for the aligned queries and further align the queries with their matched proposal. As illustrated in~\cref{fig:query_init}~(b), the start and end timestamps from the $k$-th proposal are used to initialize the boundary-level position vectors $\boldsymbol{s}^{pos}_k$ and $\boldsymbol{e}^{pos}_k$, which can also be transformed into center and duration values to initialize the instance-level position vectors $\boldsymbol{i}^{pos}_k$. At the same time, the $k$-th selected feature is employed to initialize the content vectors $\boldsymbol{s}^{con}_k$, $\boldsymbol{e}^{con}_k$, and $\boldsymbol{i}^{con}_k$ through channel splitting. This joint initialization strategy offers two benefits: (1) it further enhances the alignment between the dual-level queries, and (2) it leverages both position and semantic priors from the proposal, resulting in a better match.

\subsection{Two-Branch Decoding}
Before the two-branch decoding process, we partition the encoder feature map $X^{Enc}$ into two levels: 1) the boundary-level, which consists of the starting boundary feature $X^{Enc}_s \in \mathbb{R}^{T \times (D/4)}$ and the ending boundary feature $X^{Enc}_e \in \mathbb{R}^{T \times (D/4)}$, and 2) the instance-level feature $X^{Enc}_i\in \mathbb{R}^{T \times (D/2)}$. This partition allows the queries from each level to focus on the specific semantics relevant to their respective levels.
The decoder layers for both levels consist of a self-attention module~\cite{transformer}, a deformable cross-attention module, and a feed-forward network (FFN).

\noindent\textbf{Boundary-Level Decoding.} The boundary-level decoder layer takes boundary-level features maps $X^{Enc}_s, X^{Enc}_e$, along with the content query vectors $\boldsymbol{s}^{con}, \boldsymbol{e}^{con}$, as well as the position query vectors $\boldsymbol{s}^{pos}, \boldsymbol{e}^{pos}$ as input. After the self-attention step, we employ deformable cross-attention to attend to the proposal boundaries. Specifically, we reuse the position vector $\boldsymbol{s}^{pos}, \boldsymbol{e}^{pos}$ as reference points (as described in~\cref{eq:deform-attn}). The deformable attention attends to a small set of key sampling points around each reference point. The refinement of the content vector can be represented by:
\begin{equation}
    \begin{aligned}
        \boldsymbol{s}^{con} = \text{FFN} ( \text{DeformAttn} (\boldsymbol{s}^{con}, \boldsymbol{s}^{pos}, X^{Enc}_s) ),\\
\boldsymbol{e}^{con}= \text{FFN} ( \text{DeformAttn} (\boldsymbol{e}^{con}, \boldsymbol{e}^{pos}, X^{Enc}_e) ).
    \end{aligned}
\end{equation}
Subsequently, a regression head is applied to the refined content vectors to generate offsets $\Delta\boldsymbol{s}, \Delta\boldsymbol{e}$, which are used to refine the position vectors, as follows:
\begin{equation}
    \begin{aligned}
&\boldsymbol{s}^{pos} = \sigma(\Delta\boldsymbol{s} + \sigma^{-1}(\boldsymbol{s}^{pos})),\\
&\boldsymbol{e}^{pos} = \sigma(\Delta\boldsymbol{e} + \sigma^{-1}(\boldsymbol{e}^{pos})),
    \end{aligned}
\end{equation}
where $\sigma$ and $\sigma^{-1}$ denote the \texttt{sigmoid} and \texttt{inverse sigmoid} functions, respectively. These functions are employed to ensure that the proposal coordinates remain normalized at all times.

\noindent\textbf{Instance-Level Decoding.} Similarly to the boundary-level decoding, the instance-level decoder layer takes instance-level feature maps $X_i^{Enc}$, the content query vector $\boldsymbol{i}^{con}$, and the position query vector $\boldsymbol{i}^{pos}$ as input. After applying self-attention to model the query relationships, the content query refines itself by attending to the key semantic frames within the instance-level feature. This process utilizes the instance-level position vector as the reference point, which contains the center point and duration of the proposals. The refinement can be expressed as:
\begin{equation}
    \boldsymbol{i}^{con} = \text{FFN} ( \text{DeformAttn} (\boldsymbol{i}^{con}, \boldsymbol{i}^{pos}, X^{Enc}_i) ).
\end{equation}
Subsequently, a regression head is employed to generate offsets $\Delta i$ for refining the position vectors:
\begin{equation}
\boldsymbol{i}^{pos} = \sigma(\Delta\boldsymbol{i} + \sigma^{-1}(\boldsymbol{i}^{pos})),\\
\end{equation}

\noindent\textbf{Mutual Refinement.}
After refining the decoder queries at separate levels, we introduce a mutual refinement module to achieve complementary refinement of proposals by leveraging their matched queries. This approach allows the boundary level to benefit from the robust localization of the instance level, while the instance level can leverage the precise boundary refinement of the boundary level.
Specifically, we utilize the boundary-level position vector to refine the instance-level counterparts, which can be represented as:
\begin{equation}
\begin{aligned}
\small
& \boldsymbol{i}^{pos,0} \leftarrow \frac{\boldsymbol{i}^{pos,0} + (\boldsymbol{s}^{pos} + \boldsymbol{e}^{pos})/2}{2}, \\
& \boldsymbol{i}^{pos,1} \leftarrow \frac{\boldsymbol{i}^{pos,1} + (\boldsymbol{e}^{pos} - \boldsymbol{s}^{pos})}{2},
\end{aligned}
\end{equation}
where $\boldsymbol{i}^{pos,0}$ and $\boldsymbol{i}^{pos,1}$ represent the center point and duration contained in the instance-level position vector. Similarly, we refine the boundary level as follows:
\begin{equation}
\begin{aligned}
\small
& \boldsymbol{s}^{pos} \leftarrow \frac{\boldsymbol{s}^{pos} + (\boldsymbol{i}^{pos,0} - \boldsymbol{i}^{pos,1}/2) }{2}, \\
& \boldsymbol{e}^{pos} \leftarrow \frac{\boldsymbol{e}^{pos} + (\boldsymbol{i}^{pos,0} + \boldsymbol{i}^{pos,1}/2) }{2}.
\end{aligned}
\end{equation}
These refined position vectors are then utilized in the subsequent layers or can serve as the final predictions in the last decoder layer.

\subsection{Training}
\noindent\textbf{Label Assignment.}
The predicted action set is denoted as $\hat{\Psi}=\left\{\hat{\varphi}_n=\left(\hat{t}^{s}_{n}, \hat{t}^{e}_{n}, \hat{\boldsymbol{p}}_n\right)\right\}_{n=1}^{N_q}$, where $\hat{t}^{s}_{n}, \hat{t}^{e}_{n}$ represent the starting and ending time of predicted action instances, and $\hat{\boldsymbol{p}}_n$ represents the corresponding classification scores. The ground-truth set $\Psi$ is padded with a no-action placeholder, denoted as $\varnothing$. The cost for a  permutation $\sigma \in \mathfrak{S}_{N_q}$ of query set is defined as follows:
\begin{equation}
\begin{aligned}
\small
&\mathcal{C}_{cls}(\sigma)=\mathcal{L}_{cls}\left(\hat{\boldsymbol{p}}_{\sigma(i)},a_i\right),\\
&\mathcal{C}_{iou}(\sigma)= \mathcal{L}_{iou}\left((\hat{t}^{s}_{\sigma(i)}, \hat{t}^{e}_{\sigma(i)}), (t^{s}_{n}, t^{e}_{n})\right),\\
&\mathcal{C}_{L_1}(\sigma)= \mathcal{L}_{L_1}\left((\hat{t}^{s}_{\sigma(i)}, \hat{t}^{e}_{\sigma(i)}), (t^{s}_{n}, t^{e}_{n})\right),
\end{aligned}
\end{equation}
where $\mathcal{L}_{iou}$ represents the temporal IoU loss and $\mathcal{L}_{L1}$ denotes the $L_1$ distance. Focal loss~\cite{focal_loss} is utilized as $\mathcal{L}_{cls}$ following~\cite{deformable_detr, tadtr}. The bipartite matching between two sets aims to find the permutation $\sigma^*$ with the lowest cost:
\begin{equation}
\small
    \sigma^*=\mathop{\arg\min}\limits_{\sigma}\sum_{c_i \neq \varnothing}\alpha_{cls}\mathcal{C}_{cls}(\sigma)+\alpha_{iou}\mathcal{C}_{iou}(\sigma)+\alpha_{L_1}\mathcal{C}_{L_1}(\sigma).
\end{equation}
where $c_i$ denotes the class label. $\alpha_{cls}$, $\alpha_{iou}$, and $\alpha_{L_1}$ are the weights of each cost, set to $6$, $2$, and $5$ respectively, as in~\cite{tadtr}.

\noindent\textbf{Loss Functions.}
After obtaining the best permutation $\sigma^*$, the final optimization goal can be expressed as:
\begin{equation}
\small
\mathcal{L}=\sum_i^{N_q} \lambda_{cls} \mathcal{C}_{cls}(\sigma^*) + \mathbb{I}_{\left\{c_i \neq \varnothing\right\}}\left(\lambda_{iou}\mathcal{C}_{iou}(\sigma^*)+\lambda_{L_1}\mathcal{C}_{L_1}(\sigma^*)\right),
\end{equation}
where $\mathbb{I}(\cdot)$ is the indicator function, and $\lambda_{cls}$ and $\lambda_{iou}$ and $\lambda_{L_1}$ are set to $2$, $2$, and $5$ respectively, as in~\cite{tadtr}.

%------------------------------------------------------------------------------%
\section{Experiments}

\subsection{Dataset and Setup}
\noindent\textbf{Dataset.} We evaluate~\name~on three challenging datasets: (1) MultiTHUMOS~\cite{multi_thumos}, an extension of THUMOS14~\cite{thumos14}, containing 413 sports videos of 65 classes. The average video length is 212 seconds, and each video has an average of 97 ground-truth instances. (2) Charades~\cite{charades} comprises 9,848 videos of daily activities across 157 classes. The dataset contains an average of 6.75 action instances per video, with an average video length of 30 seconds. (3) TSU~\cite{dai2022toyota} is a dataset recorded in an indoor environment with dense annotations. Up to 5 actions can happen at the same moment. Additionally, the dataset also includes many long-term composite actions.

\begin{table}[t]
\tabstyle{3pt}
\begin{center}
\renewcommand\arraystretch{1}
\begin{tabular}{lcccccc}
\toprule
\multirow{2}{*}{Methods} &
\multirow{2}{*}{Backbone} &
\multicolumn{2}{c}{MultiTHUMOS} &
\multicolumn{2}{c}{Charades}\\
\cmidrule(lr){3-4} \cmidrule(lr){5-6} &  & Det$_\text{mAP}$ & Seg$_\text{mAP}$ & Det$_\text{mAP}$ & Seg$_\text{mAP}$\\
\midrule
R-C3D~\cite{rc3d} & C3D & -- & -- & -- & 17.6\\
Super-event~\cite{piergiovanni2018superevent} & I3D & -- & 36.4 & -- & 18.6\\
TGM~\cite{piergiovanni2019TGM} & I3D & -- & 37.2 & -- & 20.6\\
PDAN~\cite{pdan} & I3D & 17.3 & 40.2 & 8.5 & 23.7\\
Coarse-Fine~\cite{coarse_fine} & X3D & -- & -- & 6.1 & 25.1\\
MLAD~\cite{mlad} & I3D & 14.2 & 42.2 & -- & 18.4\\
CTRN~\cite{ctrn} & I3D & -- & 44.0 & -- & 25.3\\
MS-TCT~\cite{ms_tct} & I3D & 16.2 & 43.1 & 7.9 & \textbf{25.4}\\
PointTAD~\cite{pointtad} & I3D & 23.5 & 39.8 & 12.1 & 21.0\\
\midrule
\name & I3D & \textbf{32.6} & \textbf{45.5} & \textbf{15.3} & 23.2\\
\bottomrule
\end{tabular}
\end{center}
\vspace{-3mm}
\caption{\textbf{Comparison with state-of-the-art} multi-label TAD methods on MultiTHUMOS and Charades under \textbf{Det}ection-\textbf{mAP} ($\%$) and \textbf{Seg}mentation-\textbf{mAP} ($\%$).}
\vspace{-1mm}
\label{tab:cmp_multi_tad}
\end{table}

\begin{table}[t]
\tabstyle{5.5pt}
\begin{center}
\renewcommand\arraystretch{1}
\begin{tabular}{lcccc}
\toprule
Methods & Backbone & GFLOPs & Det$_\text{mAP}$ & Seg$_\text{mAP}$ \\
\midrule
R-C3D~\cite{rc3d} & C3D & -- & -- & 8.7 \\
Super-event~\cite{piergiovanni2018superevent} & I3D & 0.8 & -- & 17.2 \\
TGM~\cite{piergiovanni2019TGM} & I3D & 1.2 & -- & 26.7\\
PDAN~\cite{pdan} & I3D &  3.2 & -- & 32.7 \\
MS-TCT~\cite{ms_tct} & I3D & 6.6 & 10.6 & 33.7\\
\midrule
   \name & I3D & 5.5 & \textbf{20.8} & \textbf{34.8}\\
\bottomrule
\end{tabular}
\end{center}
\vspace{-3mm}
\caption{\textbf{Comparison with state-of-the-art} multi-label TAD methods on the TSU dataset, where the action instances are highly overlapped. The GFLOPs are presented to evaluate the computation efficiency.}
\vspace{-2mm}
\label{tab:cmp_tsu}
\end{table}

\noindent\textbf{Implementation Details.} Our method relies on offline extracted video features, and we utilize the two-stream I3D~\cite{i3d} network pre-trained on Kinetics~\cite{i3d} as the feature extractor. The video features are extracted at a stride of 4 frames, 8 frames, and 16 frames for MultiTHUMOS, Charades, and TSU, respectively. During training, for MultiTHUMOS and TSU, we crop each video feature sequence into windows of length 256 and 96, respectively, with a stride ratio of 0.75. During inference, the stride ratio is set to 0.25. For Charades, we directly input the entire video feature into the model, padding short videos with zeros for parallel computation.  We utilize AdamW~\cite{adamw} optimizer with a learning rate of 2e-4 and weight decay of 0.05. Training is performed for 30 epochs on MultiTHUMOS and Charades, and 20 epochs on TSU, with the learning rate dropping to 2e-5 during the last 3 epochs.

\noindent\textbf{Metrics.} Following~\cite{pointtad}, we evaluate our method using two metrics: detection-mAP and segmentation-mAP. Det-mAP measures boundary localization accuracy, while seg-mAP evaluates frame-wise multi-label classification precision. We report the average mAP across tIoU thresholds $[0.1: 0.1: 0.9]$ and individual mAPs at each threshold.

\begin{table}[t]
\tabstyle{5.5pt}
\begin{center}
\renewcommand\arraystretch{1}
\begin{tabular}{lccccc}
\toprule
Methods & 0.1 & 0.3 & 0.5 & Avg. \\
\midrule
% \multicolumn{2}{l}{\textbf{\textit{RGB only}}} & & & & & \\ 
BSN~\cite{BSN}+P-GCN~\cite{p_gcn} & 22.2 & 16.7 & 8.5 & 10.0\\
BSN~\cite{BSN}+ContextLoc~\cite{contextloc} & 22.9 & 18.0 & 10.8 & 11.0\\
AFSD~\cite{afsd} & 30.5 & 23.6 & 14.0 & 14.7\\
TadTR~\cite{tadtr} & 48.0 & 41.1 & 29.1 & 27.4\\
ActionFormer~\cite{actionformer}& 49.0 & 44.5 & 33.3 & 29.6 \\
TriDet~\cite{shi2023tridet} & -- & -- & 34.3 & 30.7 \\
\midrule
\name & \textbf{53.4} & \textbf{47.4} & \textbf{35.2} & \textbf{32.6}\\
\bottomrule
\end{tabular}
\end{center}
\vspace{-3mm}
\caption{\textbf{Comparison with traditional TAD methods} on MultiTHUMOS under detection-mAP ($\%$).}
\vspace{-2mm}
\label{tab:cmp_tradition_tad}
\end{table}

\subsection{Main Results}

\noindent\textbf{Comparison with State-of-The-Art Methods.}
In~\cref{tab:cmp_multi_tad}, we compare the performance of~\name~with previous multi-label TAD methods. To calculate the segmentation-mAP metric, we follow PointTAD~\cite{pointtad} by converting sparse prediction tuples into dense segmentation scores. Our~\name~outperforms all previous methods by a substantial margin under detection-mAP ($\mathbf{+9.1\%}$ on MultiTHUMOS and $\mathbf{+3.2\%}$ on Charades), highlighting its exceptional boundary localization ability. Meanwhile, even when evaluated under segmentation-mAP,~\name~delivers comparable results to methods specifically designed for the frame-wise classification task. This further emphasizes the superiority of our approach. Additionally, we present the results on the TSU dataset in~\cref{tab:cmp_tsu}, where action instances are highly overlapped.~\name~still achieves remarkable performance while maintaining favorable computational efficiency.

\noindent\textbf{Comparison with Traditional TAD Methods.} In~\cref{tab:cmp_tradition_tad}, we compare several representative methods in traditional TAD. Since MultiTHUMOS shares similar data preparation with THUMOS14, we reproduce these methods using their default hyper-parameter setting for THUMOS14, with the exception of TadTR, where the number of queries is adjusted to the same number as ours for fair comparison. While these methods demonstrate decent performance in traditional TAD, directly applying them to multi-label scenarios yields unsatisfactory results. In contrast, our~\name~takes into account the dense overlapping scenarios in our architecture design, thus achieving superior detection performance, surpassing all these methods.

\begin{figure}[t]
\centering
\includegraphics[width=0.9\linewidth]{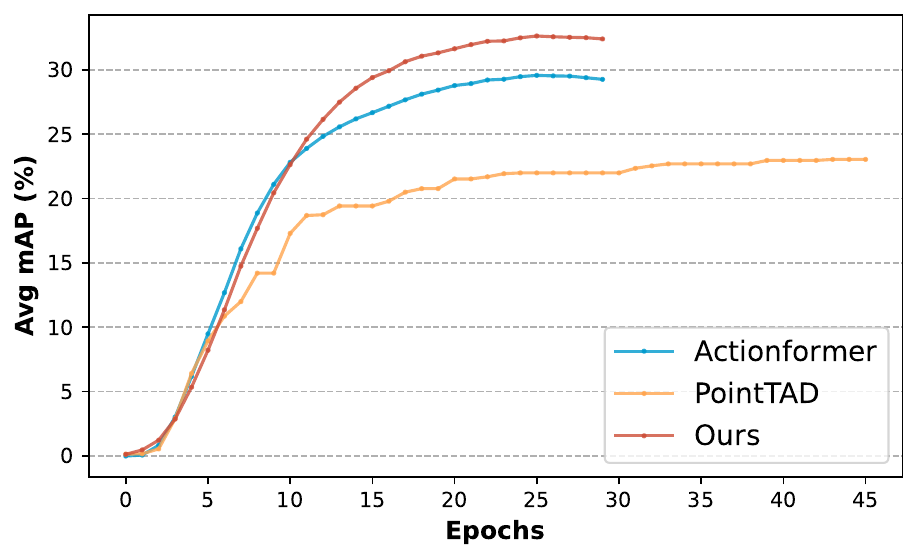}
\vspace{-2mm}
\caption{\textbf{Convergence curves} of~\name~, PointTAD~\cite{pointtad}, and ActionFormer~\cite{actionformer} on MultiTHUMOS.}
\vspace{-1mm}
\label{fig:converge}
\end{figure}

\noindent\textbf{Convergence Speed.} Query-based methods often encounter slow convergence issues~\cite{DETR,deformable_detr} compared to dense prediction methods. In~\cref{fig:converge}, we compare the convergence speeds of~\name~with PointTAD (another query-based method) and ActionFormer (dense prediction). Remarkably, our~\name~demonstrates a favorable convergence speed, thanks to the effectiveness of our two-branch collaboration structure.

\subsection{Ablation Study}
We perform ablation studies on the MultiTHUMOS dataset to evaluate the effectiveness of our proposed method and investigate alternative design choices. In the tables, the default setting is colored \colorbox{gray!30}{gray}.

\noindent\textbf{Study on Dual-Level Design.} In~\cref{tab:dual_ablation}, we present the results of our ablation study, focusing on each component of~\name. We examine the single-level results obtained by employing either the instance-level or boundary-level detection paradigm. 
The instance-level detection approach yields sub-optimal results as it lacks explicit focus on boundary information. On the other hand, the boundary-level detection approach faces challenges such as incomplete detection and difficulty in obtaining reliable scores, resulting in inferior performance.

Next, we proceed to study the effectiveness of each proposed component in a step-wise manner. We first present our baseline, which simply combines instance-level and boundary-level queries into the same detection framework. Then, we incorporate our two-branch design into this framework, enabling the decoding process to focus on specific semantics at each level. This integration leads to a promising performance gain of $\mathbf{2\%}$. Furthermore, we introduce query alignment to match the dual-level queries with the encoder proposal, enabling effective collaboration. This alignment brings an additional performance gain of $\mathbf{1.51\%}$. Lastly, the joint query initialization strategy further facilitates the alignment between queries, resulting in an additional performance gain of $\mathbf{2.09\%}$.

\begin{table}[t]
\tabstyle{2.8pt}
\begin{center}
\renewcommand\arraystretch{1}
\begin{tabular}{l c c c c c c}
\toprule
Setting & 0.1  & 0.3 & 0.5 & 0.7 & 0.9 & Avg.\\
\midrule
\textbf{\textit{Single-Level}} \\
Instance Level & 50.22 & 42.92 & 30.69 & 15.72 & 2.28 & 28.62\\
Boundary Level & 42.11 & 33.60 & 23.58 & 12.96 & 1.99 & 22.92\\
\midrule
\textbf{\textit{Dual-Level}}\\
(1):~Simple Combine & 49.10 & 40.99 & 28.11 & 14.62 & 1.65 & 27.04 \\
(2):~(1)+Two-Branch & 51.24 & 43.20 & 30.92 & 16.53 & 2.85 & 29.04\\
(3):~(2)+Query Align & 51.70 & 44.76 & 32.97 & 18.37 & 3.03 & 30.55 \\
\rowcolor{gray!30} (4):~(3)+Joint Init & \textbf{53.42} & \textbf{47.41} & \textbf{35.18} & \textbf{20.18} & \textbf{4.02} & \textbf{32.64} \\
\bottomrule
\end{tabular}
\end{center}
\vspace{-3mm}
\caption{\textbf{Ablation Study on Dual-Level}. We first show the results of single-level detection. Following that, we present our baseline, a method simply combining two-level detection. Subsequently, we perform step-wise ablations on our proposed approaches to evaluate their effectiveness.}
\label{tab:dual_ablation}
\end{table}

\begin{figure}[t]
\centering
\includegraphics[width=0.85\linewidth]{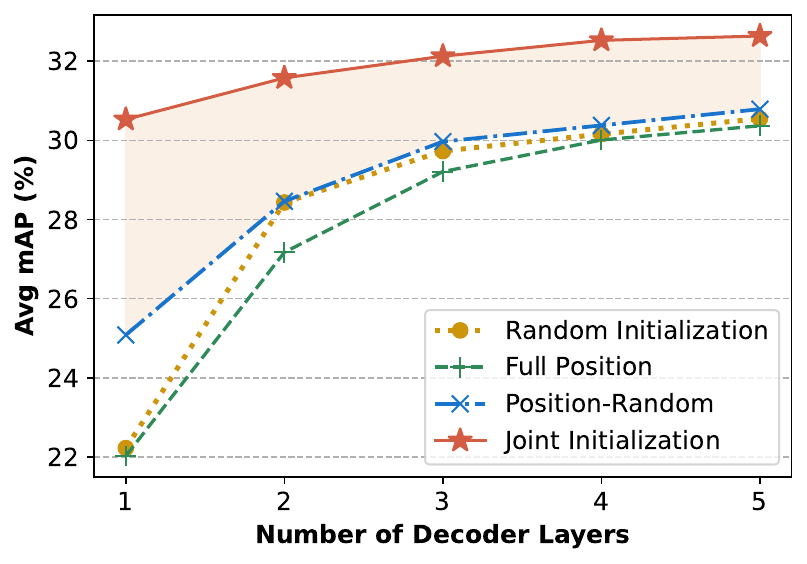}
\vspace{-1mm}
\caption{\textbf{Comparison of detection mAP at each decoder layer for different query initializations.} All initialization strategies are re-implemented in the~\name~framework. The joint initialization showcased strong detection performance from early decoding stages and continues to outperform other initialization variants as the number of decoder layers increases.}
\vspace{-2mm}
\label{fig:study_on_init}
\end{figure}

\noindent\textbf{Study on Query Initialization.} Previous query-based TAD approaches typically optimize randomly initialized queries during training to learn dataset-level action distribution. In contrast,~\name~takes advantage of position and semantic priors from the matched proposal. These priors serve two important purposes: help the decoder queries explicitly concentrate on the matched proposals, and provide an additional constraint on the aligned queries, facilitating effective collaboration. Thanks to these good qualities,~\name~enjoys prominent detection accuracy with only a few decoder layers, as depicted in~\cref{fig:study_on_init}.

Additionally, we have observed two other popular initialization methods~\cite{deformable_detr, dino} in the objection detection field.~\citet{deformable_detr} initializes both position and content vectors with proposal predictions from the encoder (denoted as ``Full Position''). However, this approach only leverages position priors for initialization, lacking crucial semantics priors necessary for fine-grained content decoding. On the other hand,~\citet{dino} goes back to learning the content vector during training while keeping the proposal predictions for position vector initialization (denoted as ``Position Random''). Although these methods achieve superior results in object detection, adapting them to a multi-level query-based framework remains non-trivial, as shown in~\cref{fig:study_on_init}.

\noindent\textbf{Alternative Choices for Mutual Refinement.} Based on the design of~\name, we explore alternative choices for the mutual refinement module in~\cref{tab:choices_for_refine}. Firstly, we consider sequential refinement, which updates the position vectors in a sequential manner. This can be done either by refining the boundary-level vectors first followed by the instance-level vectors, or vice versa. Secondly, we investigate the timing of position vector updates within the decoding process. By default, the position vectors are updated at the end of each layer. We also explore the option of updating them at the end of the entire decoding process (last layer). Furthermore, we experiment with refining the content vectors during the mutual refinement process by feeding the concatenated content vectors into a feed-forward network. Overall, our default setting benefits from parallel computation and achieves superior performance. It is also worth noting that, despite the various alternative choices explored,~\name~consistently demonstrates favorable performance, showcasing its robustness.

\begin{table}[t]
\tabstyle{2.1pt}
\begin{center}
\renewcommand\arraystretch{1}
\begin{tabular}{l c c c c c c}
\toprule
Setting & 0.1  & 0.3 & 0.5 & 0.7 & 0.9 & Avg.\\
\midrule
\textbf{\textit{Sequential Refinement}} \\
Boundary$\rightarrow$Instance & \textbf{53.76} & 47.15 & 34.84 & 19.55 & 3.75 & 32.35 \\
Instance$\rightarrow$Boundary & 53.36 & \textbf{47.44} & \textbf{35.76} & 19.60 & 3.84 & 32.50 \\
\midrule
\textbf{\textit{Parallel Refinement}} \\
\rowcolor{gray!30} Refine Position & 53.42 & 47.41 & 35.18 & \textbf{20.18} & 4.02 & \textbf{32.64} \\
Refine Pos (at last layer) & 52.51 & 46.60 & 34.71 & 18.98 & 4.00 & 31.81 \\
Refine Position\&Content & 52.54 & 46.31 & 34.23 & 19.44 & \textbf{4.13} & 31.82 \\
\bottomrule
\end{tabular}
\end{center}
\vspace{-4mm}
\caption{\textbf{Alternative design choices for mutual refinement.}}
\vspace{-2mm}
\label{tab:choices_for_refine}
\end{table}

%------------------------------------------------------------------------------%
\section{Conclusion}
In this paper, we introduce~\name, a novel dual-level query-based TAD framework.~\name~integrates both instance-level and boundary-level decoding to achieve more precise localization of temporal boundaries. To enable explicit modeling of each level's semantics, we propose a two-branch decoding structure, which allows us to capture the individual characteristics of each level.
Meanwhile, to achieve complementary refinement of action proposals, we introduce query alignment, which matches dual-level queries with encoder proposals in a one-to-one manner.
Furthermore, we propose the joint query initialization strategy that exploits rich priors from matched proposals, further enhancing the alignment.
Thanks to the dual-level design,~\name~outperforms existing TAD methods on various multi-label TAD benchmarks without the need for NMS post-processing.

\paragraph {\bf Acknowledgements.} {This work is supported by the National Key R$\&$D Program of China (No. 2022ZD0160900), the National Natural Science Foundation of China (No. 62076119, No. 61921006)), and Collaborative Innovation Center of Novel Software Technology and Industrialization.}

\appendix

% \clearpage
% \maketitlesupplementary

\begin{table*}[htbp]
\begin{center}
\resizebox{0.8\textwidth}{!}{
\begin{tabular}{lc|cccccc|cccc}
\toprule
\multirow{2.4}{*}{\textbf{Method}} &
\multirow{2.4}{*}{\textbf{Backbone}} &
\multicolumn{6}{c|}{\textbf{THUMOS14~\cite{thumos14}}} & \multicolumn{4}{c}{\textbf{ActivityNet-v1.3~\cite{caba2015activitynet}}}\\
\cmidrule(lr){3-8} \cmidrule(lr){9-12}
& & \textbf{0.3} & \textbf{0.4} & \textbf{0.5} & \textbf{0.6} & \textbf{0.7} & \textbf{Avg.} & \textbf{0.5} & \textbf{0.75} & \textbf{0.95} & \textbf{Avg.}\\
\midrule
\rowcolor{gray!20} \multicolumn{12}{l}{\textbf{\textit{Standard Methods}}}\\
BMN~\cite{bmn} & TSN~\cite{tsn} & 56.0 & 47.4 & 38.8 & 29.7 & 20.5 & 38.5 & 50.1 & 34.8 & 8.3 & 33.9 \\
G-TAD~\cite{g_tad} & TSN~\cite{tsn} & 54.5 & 47.6 & 40.2 & 30.8 & 23.4 & 39.3 & 50.4 & 34.6 & 9.0 & 34.1 \\
BC-GNN~\cite{bai2020boundary} & TSN~\cite{tsn} & 57.1 & 49.1 & 40.4 & 31.2 & 23.1 & 40.2 & 50.6 & 34.8 & 9.4 & 34.3 \\
TAL-MR~\cite{zhao2020bottom} & I3D~\cite{i3d} & 53.9 & 50.7 & 45.4 & 38.0 & 28.5 & 43.3 & 43.5 & 33.9 & 9.2 & 30.2 \\
TCA-Net~\cite{tcanet} & TSN~\cite{tsn} & 60.6 & 53.2 & 44.6 & 36.8 & 26.7 & 44.3 & 52.3 & 36.7 & 6.9 & 35.5 \\
BMN-CSA~\cite{bai2020boundary} & TSN~\cite{tsn} & 64.4 & 58.0 & 49.2 & 38.2 & 27.8 & 47.7 & 52.4 & 36.2 & 5.2 & 35.4 \\
VSGN~\cite{vsgn} & TSN~\cite{tsn} & 66.7 & 60.4 & 52.4 & 41.0 & 30.4 & 50.2 & 52.4 & 36.0 & 8.4 & 35.1 \\
ContextLoc~\cite{contextloc} & I3D~\cite{i3d} & 68.3 & 63.8 & 54.3 & 41.8 & 26.2 & 50.9 & 56.0 & 35.2 & 3.6 & 34.2 \\
RCL~\cite{wang2022rcl} & I3D~\cite{i3d} & 70.1 & 62.3 & 52.9 & 42.7 & 30.7 & 51.0 & 51.7 & 35.3 & 8.0 & 34.4 \\
AFSD~\cite{afsd} & I3D~\cite{i3d} & 67.3 & 62.4 & 55.5 & 43.7 & 31.1 & 52.0 & 52.4 & 35.3 & 6.5 & 34.4 \\
DCAN~\cite{dcan} & TSN~\cite{tsn} & 68.2 & 62.7 & 54.1 & 43.9 & 32.6 & 52.3 & 51.8 & 36.0 & 9.5 & 35.4 \\
TAGS~\cite{nag2022proposal} & I3D~\cite{i3d} & 68.6 & 63.8 & 57.0 & 46.3 & 31.8 & 52.8 & 56.3 & \textbf{36.8} & \textbf{9.6} & \textbf{36.5} \\
MUSES~\cite{liu2021multi} & I3D~\cite{i3d} & 68.9 & 64.0 & 56.9 & 46.3 & 31.0 & 53.4 & 50.0 & 35.0 & 6.6 & 34.0 \\
\citet{zhu2022learning} & I3D~\cite{i3d} & 72.1 & 65.9 & 57.0 & 44.2 & 28.5 & 53.5 & \textbf{58.1} & 36.3 & 6.2 & 35.2 \\
ActionFormer~\cite{actionformer} & I3D~\cite{i3d} & 82.1 & 77.8 & 71.0 & 59.4 & 43.9 & 66.8 & 53.5 & 36.2 & 8.2 & 35.6 \\
TriDet~\cite{shi2023tridet} & I3D~\cite{i3d} & \textbf{83.6} & \textbf{80.1} & \textbf{72.9} & \textbf{62.4} & \textbf{47.4} & \textbf{69.3} &  -- & -- & -- & --\\
\midrule
\rowcolor{gray!20} \multicolumn{12}{l}{\textbf{\textit{Query-Based Methods}}}\\
TadTR~\cite{tadtr} & I3D~\cite{i3d} & 62.4 & 57.4 & 49.2 & 37.8 & 26.3 & 46.6 & 49.1 & 32.6 & 8.5 & 32.3 \\
RTD-Net~\cite{rtdnet} & I3D~\cite{i3d} & 68.3 & 62.3 & 51.9 & 38.8 & 23.7 & 49.0 & 47.2 & 30.7 & \textbf{8.6} & 30.8 \\
DINO~\cite{dino} & I3D~\cite{i3d} & 69.8 & 63.1 & 53.7 & 41.5 & 26.4 & 50.9 & -- & -- & -- & --\\
ReAct~\cite{react} & TSN~\cite{tsn} & 69.2 & 65.0 & 57.1 & 47.8 & 35.6 & 55.0 & 49.6 & 33.0 & 8.6 & 32.6 \\ 
Self-DETR~\cite{self_detr} & I3D~\cite{i3d} & 74.6 & 69.5 & 60.0 & 47.6 & 31.8 & 56.7 & 52.3 & 33.7 & 8.4 & 33.8\\
\textbf{DualDETR} & I3D~\cite{i3d} & \textbf{82.9} & \textbf{78.0} & \textbf{70.4} & \textbf{58.5} & \textbf{44.4} & \textbf{66.8} & \textbf{52.6} & \textbf{35.0} & 7.8 & \textbf{34.3} \\
\bottomrule
\end{tabular}}
\end{center}
\vspace{-4mm}
\caption{\textbf{DualDTER's performance on THUMOS14 and ActivityNet1.3.} The results of other methods are mainly from Self-DETR~\cite{self_detr}.}
\label{tab:traditional_tad_dataset}
\end{table*}

%------------------------------------------------------------------------------%
\section{More Details}

\noindent\textbf{Detection Head.} Following previous studies~\cite{DETR, deformable_detr, rtdnet, pointtad}, we apply a linear projection to the instance-level content vectors $\boldsymbol{i}^{con}$ to generate the classification score:
\begin{equation}
    \hat{\boldsymbol{p}} = \text{Linear} (\boldsymbol{i}^{con}).
\end{equation}
The classification score, $\hat{\boldsymbol{p}}$, will be used in three scenarios: 1) selecting encoder proposals in the query alignment strategy, 2) performing bipartite matching for assigning ground truth, and 3) calculating the classification loss. Moreover, we employ a Multi-Layer Perception (MLP) with ReLU activation to generate proposal offsets. Specifically, the boundary-level content vectors $\boldsymbol{s}^{con}, \boldsymbol{e}^{con}$ are used to compute boundary-level offsets, while the instance-level content vectors $\boldsymbol{i}^{con}$ are utilized to generate instance-level offsets:
\begin{equation}
    \begin{aligned}
    \Delta \boldsymbol{s} = \text{MLP} (\boldsymbol{s}^{con}),\\
    \Delta \boldsymbol{e} = \text{MLP} (\boldsymbol{e}^{con}),\\
    \Delta \boldsymbol{i} = \text{MLP} (\boldsymbol{i}^{con}).
    \end{aligned}
\end{equation}
These offsets are subsequently employed to refine their respective position vectors.

The detection head is appended to the final encoder layer as well as each decoder layer. Detection losses are computed at these stages to optimize the model. Furthermore, to preserve the alignment between dual-level queries, we share ground truth obtained through bipartite matching among each aligned query.

\noindent\textbf{Training Details.} 
DualDETR is trained on two NVIDIA TITAN Xp GPUs, with a batch size of 16 per GPU. To ensure stable training, we employ ModelEMA~\cite{modelema} and gradient clipping following~\cite{actionformer}. The random seed is fixed at 42 to ensure reproducibility.

\section{Additional Experiments}

\noindent\textbf{Traditional TAD Benchmarks.} 
The performance of DualDETR on traditional benchmarks, THUMOS14~\cite{thumos14} and AcitvityNet1.3~\cite{caba2015activitynet}, is presented in~\cref{tab:traditional_tad_dataset}. DualDETR surpasses previous query-based methods by a significant margin at all IoU thresholds on THUMOS14, achieving an impressive average mAP gain of $\mathbf{10.1\%}$. When compared to standard methods that rely on NMS post-processing, DualDETR exhibits comparable performance to the state-of-the-art method ActionFormer~\cite{actionformer}. Moreover, on ActivityNet1.3, DualDETR also outperforms all previous query-based methods. These results further demonstrate DualDETR's superiority in action detection tasks.

\noindent\textbf{Study on Number of Queries.} In~\cref{tab:ablation_num_query}, we analyze the effectiveness of the number of decoder queries. We find that the optimal number of queries is 150, 25, and 96 for MultiTHUMOS, Charades, and TSU, respectively. This observation aligns with the number of ground truth instances per video in each dataset, which is approximately 97, 6.8, and 77 for MultiTHUMOS, Charades, and TSU, respectively.

\begin{table}[htbp]
\tabstyle{3pt}
\begin{center}
\begin{tabular}{l | c c c c c}
\toprule
\rowcolor{gray!10} \# Queries ($N_q$) & 80  & 120 & 150 & 180 & 250 \\
MultiTHUMOS~\cite{multi_thumos} & 32.33 & 32.50 & \textbf{32.64} & 32.60 & 32.63 \\
\midrule
\rowcolor{gray!10} \# Queries ($N_q$) & 10  & 25 & 40 & 55 & 70 \\
Charades~\cite{charades} & 14.55 & \textbf{15.62} & 15.27 & 14.26 & 13.58 \\
\midrule
\rowcolor{gray!10} \# Queries ($N_q$) & 20  & 40 & 60 & 80 & 96 \\
TSU~\cite{dai2022toyota} & 18.56 & 20.19 & 20.59 & 20.73 & \textbf{20.81} \\
\bottomrule
\end{tabular}
\end{center}
\vspace{-4mm}
\caption{\textbf{Ablation study on the number of decoder queries.}}
\label{tab:ablation_num_query}
\end{table}

\noindent\textbf{Study on Number of Layers.} 
In~\cref{tab:ablation_num_layer}, we examine the impact of the number of encoder and decoder layers on the MultiTHUMOS dataset. Our default configuration includes 6 encoder layers and 5 decoder layers. Thanks to the joint initialization strategy, the performance remains consistently strong even with a reduced number of decoder layers. In terms of average performance, our default setting proves to be the most effective.

\begin{table}[htbp]
\tabstyle{4.8pt}
\begin{center}
\begin{tabular}{c c | c c c c c | c}
\toprule
$L_E$ & $L_D$ & 0.1 & 0.3 & 0.5 & 0.7 & 0.9 & Avg. \\
\midrule
5 & 5 & 51.90 & 45.52 & 33.54 & 19.02 & 3.83 & 31.25\\
6 & 3 & 52.69 & 46.66 & 34.52 & 19.47 & 3.75 & 31.93 \\
6 & 4 & 53.39 & \textbf{47.42} & \textbf{35.19} & 19.93 & 3.88 & 32.52 \\
6 & 5 & \textbf{53.42} & 47.41 & 35.18 & \textbf{20.18} & 4.02 & \textbf{32.64} \\
6 & 6 & 52.95 & 46.30 & 34.06 & 19.47 & \textbf{4.30} & 31.90 \\
\bottomrule
\end{tabular}
\end{center}
\vspace{-4mm}
\caption{\textbf{Ablation study on the number of encoder and decoder layers on MultiTHUMOS.}}
\label{tab:ablation_num_layer}
\end{table}

\noindent\textbf{Inference Efficiency.} 
We report the efficiency comparison on multiTHUMOS with two competitive methods ActionFormer~\cite{actionformer} and TriDet~\cite{shi2023tridet} in~\cref{tab:inference_efficiency}. DualDETR achieves the highest mAP with the least latency among all methods, perfectly balancing the efficiency-performance tradeoff. The suffix of DualDETR in the table indicates the number of decoder queries employed.

\begin{table}[h]
\begin{center}
\scalebox{0.78}
{\begin{tabular}{c|ccccc}
\toprule
 \textbf{Method} & \textbf{GPU} & \textbf{Param}  & \textbf{GMACs} & \textbf{Latency} & \textbf{mAP}\\
 \midrule
 ActionFormer & A100 & 27.90M & 45.3 & 224ms & 29.6\\
 TriDet & A100 & 15.25M & 43.7 & 167ms & 30.7\\\
 \textbf{DualDETR}$_{\text{q80}}$ & TITAN Xp & 21.77M & 66.3 & 65ms & 32.3\\
 \textbf{DualDETR}$_{\text{q150}}$ & TITAN Xp & 21.77M & 80.3 & 69ms & 32.6\\
\bottomrule
\end{tabular}}
\end{center}
\vspace{-4mm}
\caption{\textbf{Inference efficiency.}}
\label{tab:inference_efficiency}
\end{table}

\begin{figure*}[t]
\centering
\includegraphics[width=0.88\linewidth]{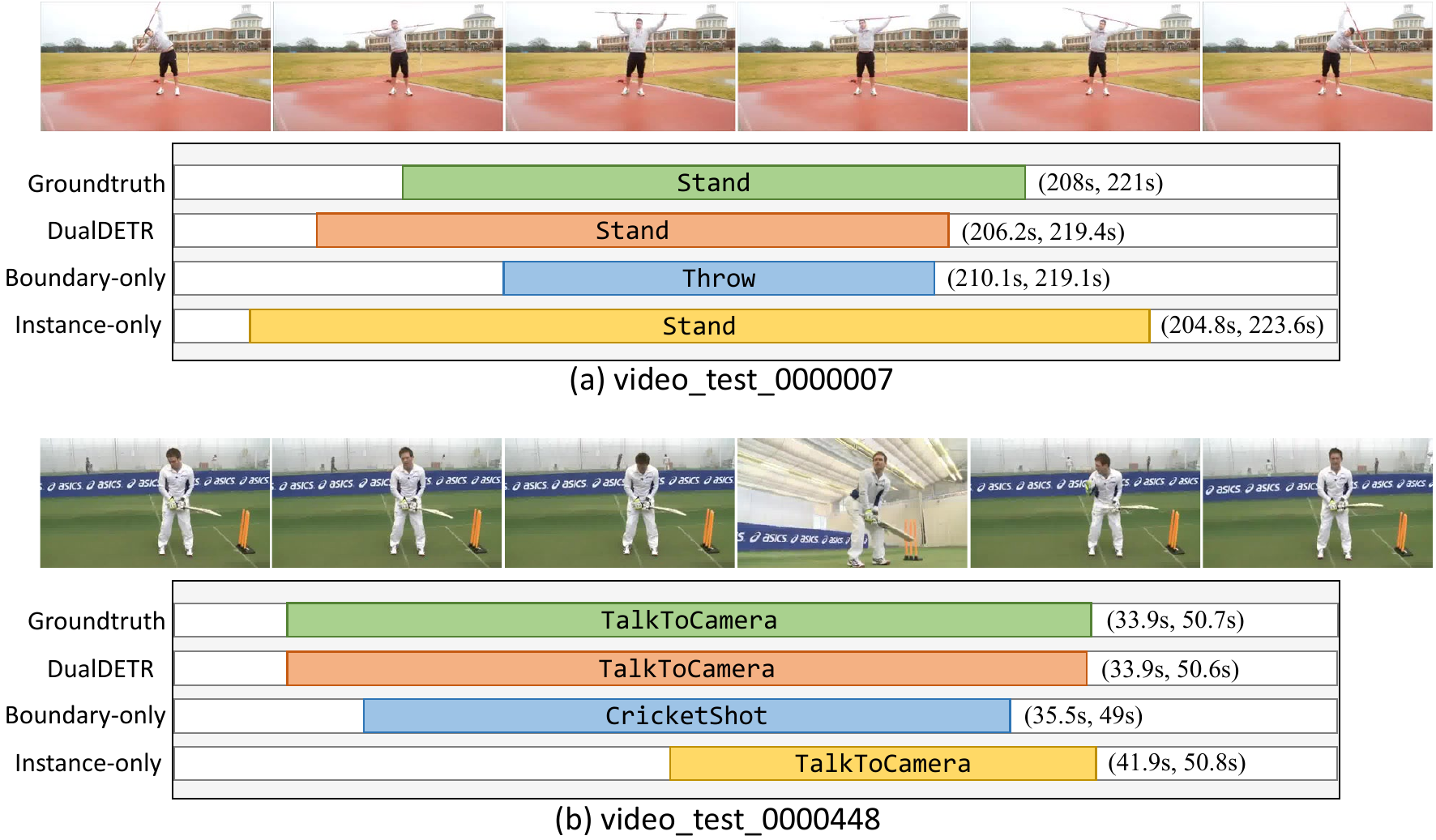}
\vspace{-2mm}
\caption{\textbf{Qualitative comparison} between predictions of single-level detection (boundary-level only and instance-level only) and our dual-level detection. The videos are from the MultiTHUMOS dataset.}
\label{fig:different}
\end{figure*}

\begin{figure*}[t]
\centering
\includegraphics[width=0.88\linewidth]{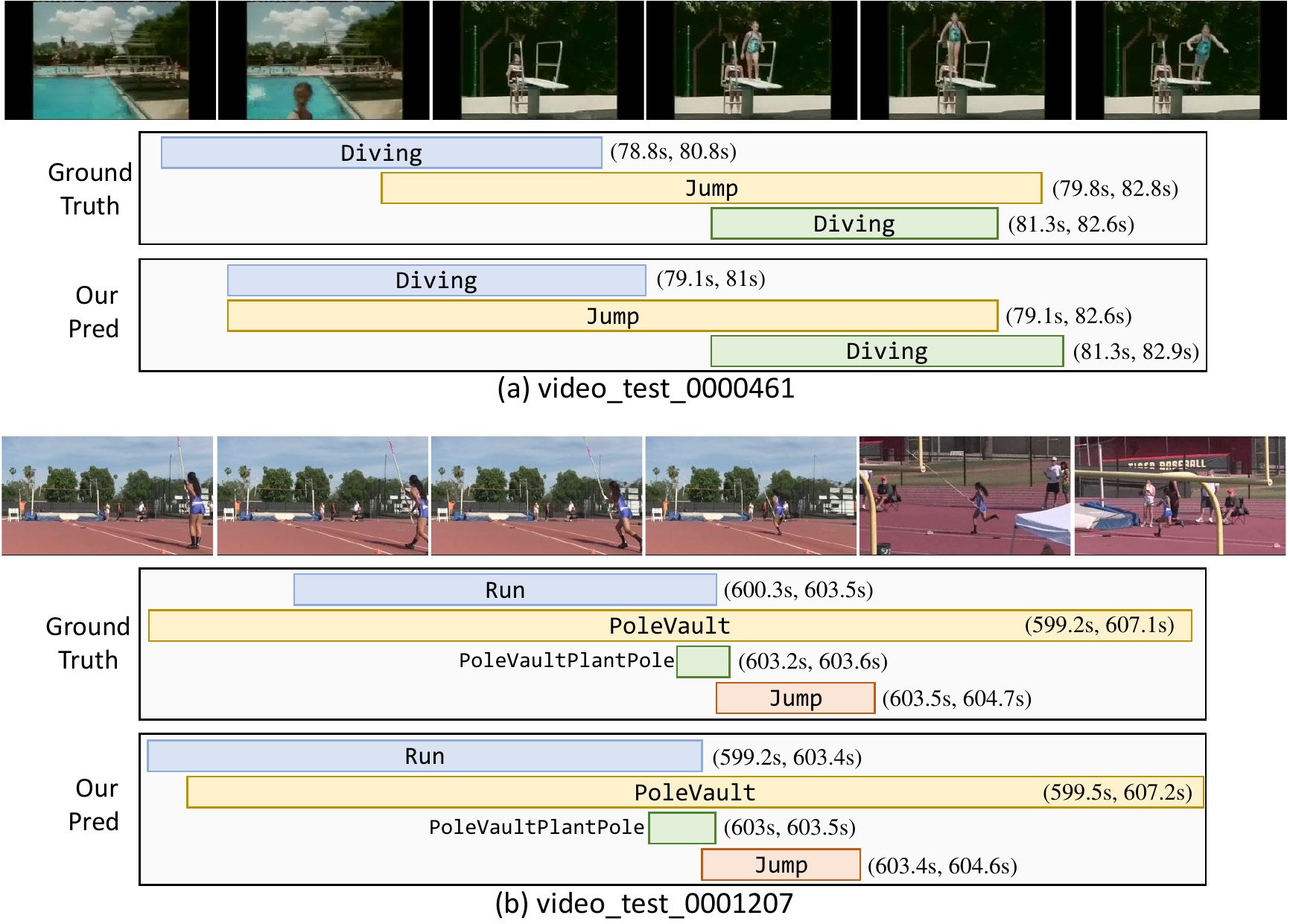}
\vspace{-2mm}
\caption{\textbf{Qualitative results} under high-overlap scenarios. The videos are from the MultiTHUMOS dataset.}
\label{fig:overlap}
\end{figure*}

\noindent\textbf{Qualitative Results.}
To further compare different detection paradigms, we present qualitative results in~\cref{fig:different}. Boundary-level detection demonstrates high accuracy in boundary detection but lacks reliable semantic labels. On the other hand, instance-level detection achieves robust detection but sub-optimal boundary localization. Our proposed DualDETR combines both paradigms effectively, offering reliable recognition and precise boundary localization simultaneously. Additionally, we provide qualitative results for high-overlap action regions in~\cref{fig:overlap}. Our method excels in handling complex situations, showcasing the strong applicability of DualDETR in multi-label action detection scenarios.

\section{Limitation and Future Work} 
In the query alignment strategy, where each query is matched with an encoder proposal, the maximum number of queries is constrained by the number of features in the encoder feature map. If situations arise where a larger number of queries is required, additional modules must be devised. Furthermore, to maintain efficiency during training and testing, DualDETR operates on pre-extracted video features following previous practice, which overlooks the gap between pre-training and downstream tasks. However, with the emergence of parameter-efficient fine-tuning techniques like LoRA~\cite{lora}, Adapter~\cite{adapter}, and Prompt Tuning~\cite{xu2023progressive, xu2023dpl}, there is a growing opportunity to explore efficient end-to-end approaches for action detection tasks.

{
    \small
    \bibliographystyle{ieeenat_fullname}
    \bibliography{main}

\begin{thebibliography}{79}
\providecommand{\natexlab}[1]{#1}
\providecommand{\url}[1]{\texttt{#1}}
\expandafter\ifx\csname urlstyle\endcsname\relax
  \providecommand{\doi}[1]{doi: #1}\else
  \providecommand{\doi}{doi: \begingroup \urlstyle{rm}\Url}\fi

\bibitem[Alyahya et~al.(2022)Alyahya, Alghannam, and Alhussan]{9857383}
Munirah Alyahya, Shahad Alghannam, and Taghreed Alhussan.
\newblock Temporal driver action localization using action classification methods.
\newblock In \emph{2022 IEEE/CVF Conference on Computer Vision and Pattern Recognition Workshops (CVPRW)}, pages 3318--3325, 2022.

\bibitem[Bai et~al.(2020)Bai, Wang, Tong, Yang, Liu, and Liu]{bai2020boundary}
Yueran Bai, Yingying Wang, Yunhai Tong, Yang Yang, Qiyue Liu, and Junhui Liu.
\newblock Boundary content graph neural network for temporal action proposal generation.
\newblock In \emph{Computer Vision--ECCV 2020: 16th European Conference, Glasgow, UK, August 23--28, 2020, Proceedings, Part XXVIII 16}, pages 121--137. Springer, 2020.

\bibitem[Caba~Heilbron et~al.(2015)Caba~Heilbron, Escorcia, Ghanem, and Carlos~Niebles]{caba2015activitynet}
Fabian Caba~Heilbron, Victor Escorcia, Bernard Ghanem, and Juan Carlos~Niebles.
\newblock Activitynet: A large-scale video benchmark for human activity understanding.
\newblock In \emph{Proceedings of the ieee conference on computer vision and pattern recognition}, pages 961--970, 2015.

\bibitem[Carion et~al.(2020)Carion, Massa, Synnaeve, Usunier, Kirillov, and Zagoruyko]{DETR}
Nicolas Carion, Francisco Massa, Gabriel Synnaeve, Nicolas Usunier, Alexander Kirillov, and Sergey Zagoruyko.
\newblock End-to-end object detection with transformers.
\newblock In \emph{European conference on computer vision}, pages 213--229. Springer, 2020.

\bibitem[Carreira and Zisserman(2017)]{i3d}
Joao Carreira and Andrew Zisserman.
\newblock Quo vadis, action recognition? a new model and the kinetics dataset.
\newblock In \emph{proceedings of the IEEE Conference on Computer Vision and Pattern Recognition}, pages 6299--6308, 2017.

\bibitem[Chao et~al.(2018)Chao, Vijayanarasimhan, Seybold, Ross, Deng, and Sukthankar]{TAL_Net}
Yu-Wei Chao, Sudheendra Vijayanarasimhan, Bryan Seybold, David~A Ross, Jia Deng, and Rahul Sukthankar.
\newblock Rethinking the faster r-cnn architecture for temporal action localization.
\newblock In \emph{Proceedings of the IEEE conference on computer vision and pattern recognition}, pages 1130--1139, 2018.

\bibitem[Chen et~al.(2022{\natexlab{a}})Chen, Zheng, Wang, and Lu]{dcan}
Guo Chen, Yin-Dong Zheng, Limin Wang, and Tong Lu.
\newblock Dcan: Improving temporal action detection via dual context aggregation.
\newblock In \emph{Proceedings of the AAAI conference on artificial intelligence}, pages 248--257, 2022{\natexlab{a}}.

\bibitem[Chen et~al.(2022{\natexlab{b}})Chen, Wei, Zeng, and Wang]{conditional_detr_v2}
Xiaokang Chen, Fangyun Wei, Gang Zeng, and Jingdong Wang.
\newblock Conditional detr v2: Efficient detection transformer with box queries.
\newblock \emph{arXiv preprint arXiv:2207.08914}, 2022{\natexlab{b}}.

\bibitem[Dai et~al.(2021{\natexlab{a}})Dai, Das, and Bremond]{ctrn}
Rui Dai, Srijan Das, and Francois Bremond.
\newblock Ctrn: Class-temporal relational network for action detection.
\newblock \emph{arXiv preprint arXiv:2110.13473}, 2021{\natexlab{a}}.

\bibitem[Dai et~al.(2021{\natexlab{b}})Dai, Das, Minciullo, Garattoni, Francesca, and Bremond]{pdan}
Rui Dai, Srijan Das, Luca Minciullo, Lorenzo Garattoni, Gianpiero Francesca, and Fran{\c{c}}ois Bremond.
\newblock Pdan: Pyramid dilated attention network for action detection.
\newblock In \emph{Proceedings of the IEEE/CVF Winter Conference on Applications of Computer Vision}, pages 2970--2979, 2021{\natexlab{b}}.

\bibitem[Dai et~al.(2022{\natexlab{a}})Dai, Das, Kahatapitiya, Ryoo, and Br{\'e}mond]{ms_tct}
Rui Dai, Srijan Das, Kumara Kahatapitiya, Michael~S Ryoo, and Fran{\c{c}}ois Br{\'e}mond.
\newblock Ms-tct: multi-scale temporal convtransformer for action detection.
\newblock In \emph{Proceedings of the IEEE/CVF Conference on Computer Vision and Pattern Recognition}, pages 20041--20051, 2022{\natexlab{a}}.

\bibitem[Dai et~al.(2022{\natexlab{b}})Dai, Das, Sharma, Minciullo, Garattoni, Bremond, and Francesca]{dai2022toyota}
Rui Dai, Srijan Das, Saurav Sharma, Luca Minciullo, Lorenzo Garattoni, Francois Bremond, and Gianpiero Francesca.
\newblock Toyota smarthome untrimmed: Real-world untrimmed videos for activity detection.
\newblock \emph{IEEE Transactions on Pattern Analysis and Machine Intelligence}, 45\penalty0 (2):\penalty0 2533--2550, 2022{\natexlab{b}}.

\bibitem[Ding et~al.(2022{\natexlab{a}})Ding, Li, Yang, Qian, Xu, Chen, Wang, and Xiong]{ding2022motion}
Shuangrui Ding, Maomao Li, Tianyu Yang, Rui Qian, Haohang Xu, Qingyi Chen, Jue Wang, and Hongkai Xiong.
\newblock Motion-aware contrastive video representation learning via foreground-background merging.
\newblock In \emph{Proceedings of the IEEE/CVF Conference on Computer Vision and Pattern Recognition}, pages 9716--9726, 2022{\natexlab{a}}.

\bibitem[Ding et~al.(2022{\natexlab{b}})Ding, Qian, and Xiong]{ding2022dual}
Shuangrui Ding, Rui Qian, and Hongkai Xiong.
\newblock Dual contrastive learning for spatio-temporal representation.
\newblock In \emph{Proceedings of the 30th ACM international conference on multimedia}, pages 5649--5658, 2022{\natexlab{b}}.

\bibitem[Ding et~al.(2023)Ding, Zhao, Zhang, Qian, Xiong, and Tian]{ding2023prune}
Shuangrui Ding, Peisen Zhao, Xiaopeng Zhang, Rui Qian, Hongkai Xiong, and Qi Tian.
\newblock Prune spatio-temporal tokens by semantic-aware temporal accumulation.
\newblock In \emph{Proceedings of the IEEE/CVF International Conference on Computer Vision}, pages 16945--16956, 2023.

\bibitem[Gao et~al.(2020)Gao, Shi, Wang, Li, Yuan, Ge, and Zhou]{rapnet}
Jialin Gao, Zhixiang Shi, Guanshuo Wang, Jiani Li, Yufeng Yuan, Shiming Ge, and Xi Zhou.
\newblock Accurate temporal action proposal generation with relation-aware pyramid network.
\newblock In \emph{Proceedings of the AAAI conference on artificial intelligence}, pages 10810--10817, 2020.

\bibitem[Gao et~al.(2022)Gao, Wang, Han, and Guo]{adamixer}
Ziteng Gao, Limin Wang, Bing Han, and Sheng Guo.
\newblock Adamixer: {A} fast-converging query-based object detector.
\newblock In \emph{Proceedings of the IEEE/CVF Conference on Computer Vision and Pattern Recognition}, pages 5354--5363, 2022.

\bibitem[Gao et~al.(2023)Gao, Qiao, and Dou]{gao2023haan}
Zikai Gao, Peng Qiao, and Yong Dou.
\newblock Haan: Human action aware network for multi-label temporal action detection.
\newblock In \emph{Proceedings of the 31st ACM International Conference on Multimedia}, pages 5059--5069, 2023.

\bibitem[Giancola et~al.(2018)Giancola, Amine, Dghaily, and Ghanem]{giancola2018soccernet}
Silvio Giancola, Mohieddine Amine, Tarek Dghaily, and Bernard Ghanem.
\newblock Soccernet: A scalable dataset for action spotting in soccer videos.
\newblock In \emph{Proceedings of the IEEE conference on computer vision and pattern recognition workshops}, pages 1711--1721, 2018.

\bibitem[He et~al.(2017)He, Gkioxari, Doll{\'a}r, and Girshick]{mask_rcnn}
Kaiming He, Georgia Gkioxari, Piotr Doll{\'a}r, and Ross Girshick.
\newblock Mask r-cnn.
\newblock In \emph{Proceedings of the IEEE international conference on computer vision}, pages 2961--2969, 2017.

\bibitem[Houlsby et~al.(2019)Houlsby, Giurgiu, Jastrzebski, Morrone, De~Laroussilhe, Gesmundo, Attariyan, and Gelly]{adapter}
Neil Houlsby, Andrei Giurgiu, Stanislaw Jastrzebski, Bruna Morrone, Quentin De~Laroussilhe, Andrea Gesmundo, Mona Attariyan, and Sylvain Gelly.
\newblock Parameter-efficient transfer learning for nlp.
\newblock In \emph{International Conference on Machine Learning}, pages 2790--2799. PMLR, 2019.

\bibitem[Hu et~al.(2021)Hu, Shen, Wallis, Allen-Zhu, Li, Wang, Wang, and Chen]{lora}
Edward~J Hu, Yelong Shen, Phillip Wallis, Zeyuan Allen-Zhu, Yuanzhi Li, Shean Wang, Lu Wang, and Weizhu Chen.
\newblock Lora: Low-rank adaptation of large language models.
\newblock \emph{arXiv preprint arXiv:2106.09685}, 2021.

\bibitem[Huang et~al.(2017)Huang, Li, Pleiss, Liu, Hopcroft, and Weinberger]{modelema}
Gao Huang, Yixuan Li, Geoff Pleiss, Zhuang Liu, John~E Hopcroft, and Kilian~Q Weinberger.
\newblock Snapshot ensembles: Train 1, get m for free.
\newblock \emph{arXiv preprint arXiv:1704.00109}, 2017.

\bibitem[Huang et~al.(2020)Huang, Xiong, Rao, Wang, and Lin]{huang2020movienet}
Qingqiu Huang, Yu Xiong, Anyi Rao, Jiaze Wang, and Dahua Lin.
\newblock Movienet: A holistic dataset for movie understanding.
\newblock In \emph{European Conference on Computer Vision}, pages 709--727. Springer, 2020.

\bibitem[Idrees et~al.(2017)Idrees, Zamir, Jiang, Gorban, Laptev, Sukthankar, and Shah]{thumos14}
Haroon Idrees, Amir~R Zamir, Yu-Gang Jiang, Alex Gorban, Ivan Laptev, Rahul Sukthankar, and Mubarak Shah.
\newblock The thumos challenge on action recognition for videos “in the wild”.
\newblock \emph{Computer Vision and Image Understanding}, 155:\penalty0 1--23, 2017.

\bibitem[Kahatapitiya and Ryoo(2021)]{coarse_fine}
Kumara Kahatapitiya and Michael~S Ryoo.
\newblock Coarse-fine networks for temporal activity detection in videos.
\newblock In \emph{Proceedings of the IEEE/CVF Conference on Computer Vision and Pattern Recognition}, pages 8385--8394, 2021.

\bibitem[Kim et~al.(2023)Kim, Lee, and Heo]{self_detr}
Jihwan Kim, Miso Lee, and Jae-Pil Heo.
\newblock Self-feedback detr for temporal action detection.
\newblock In \emph{Proceedings of the IEEE/CVF International Conference on Computer Vision}, pages 10286--10296, 2023.

\bibitem[Li et~al.(2020)Li, Wang, Wang, and Wu]{li2020actions}
Yixuan Li, Zixu Wang, Limin Wang, and Gangshan Wu.
\newblock Actions as moving points.
\newblock In \emph{Computer Vision--ECCV 2020: 16th European Conference}, pages 68--84. Springer, 2020.

\bibitem[Li et~al.(2021)Li, Chen, He, Wang, Wu, and Wang]{Li_2021_ICCV}
Yixuan Li, Lei Chen, Runyu He, Zhenzhi Wang, Gangshan Wu, and Limin Wang.
\newblock Multisports: A multi-person video dataset of spatio-temporally localized sports actions.
\newblock In \emph{Proceedings of the IEEE/CVF International Conference on Computer Vision (ICCV)}, pages 13536--13545, 2021.

\bibitem[Lin et~al.(2020)Lin, Li, Wang, Tai, Luo, Cui, Wang, Li, Huang, and Ji]{dbg}
Chuming Lin, Jian Li, Yabiao Wang, Ying Tai, Donghao Luo, Zhipeng Cui, Chengjie Wang, Jilin Li, Feiyue Huang, and Rongrong Ji.
\newblock Fast learning of temporal action proposal via dense boundary generator.
\newblock In \emph{Proceedings of the AAAI conference on artificial intelligence}, pages 11499--11506, 2020.

\bibitem[Lin et~al.(2021)Lin, Xu, Luo, Wang, Tai, Wang, Li, Huang, and Fu]{afsd}
Chuming Lin, Chengming Xu, Donghao Luo, Yabiao Wang, Ying Tai, Chengjie Wang, Jilin Li, Feiyue Huang, and Yanwei Fu.
\newblock Learning salient boundary feature for anchor-free temporal action localization.
\newblock In \emph{Proceedings of the IEEE/CVF Conference on Computer Vision and Pattern Recognition}, pages 3320--3329, 2021.

\bibitem[Lin et~al.(2018)Lin, Zhao, Su, Wang, and Yang]{BSN}
Tianwei Lin, Xu Zhao, Haisheng Su, Chongjing Wang, and Ming Yang.
\newblock Bsn: Boundary sensitive network for temporal action proposal generation.
\newblock In \emph{Proceedings of the European conference on computer vision (ECCV)}, pages 3--19, 2018.

\bibitem[Lin et~al.(2019)Lin, Liu, Li, Ding, and Wen]{bmn}
Tianwei Lin, Xiao Liu, Xin Li, Errui Ding, and Shilei Wen.
\newblock Bmn: Boundary-matching network for temporal action proposal generation.
\newblock In \emph{Proceedings of the IEEE/CVF international conference on computer vision}, pages 3889--3898, 2019.

\bibitem[Lin et~al.(2017)Lin, Goyal, Girshick, He, and Doll{\'a}r]{focal_loss}
Tsung-Yi Lin, Priya Goyal, Ross Girshick, Kaiming He, and Piotr Doll{\'a}r.
\newblock Focal loss for dense object detection.
\newblock In \emph{Proceedings of the IEEE international conference on computer vision}, pages 2980--2988, 2017.

\bibitem[Liu et~al.(2020)Liu, Zhao, Su, and Hu]{tsi}
Shuming Liu, Xu Zhao, Haisheng Su, and Zhilan Hu.
\newblock Tsi: Temporal scale invariant network for action proposal generation.
\newblock In \emph{Proceedings of the Asian Conference on Computer Vision}, 2020.

\bibitem[Liu et~al.(2022{\natexlab{a}})Liu, Li, Zhang, Yang, Qi, Su, Zhu, and Zhang]{dab_detr}
Shilong Liu, Feng Li, Hao Zhang, Xiao Yang, Xianbiao Qi, Hang Su, Jun Zhu, and Lei Zhang.
\newblock Dab-detr: Dynamic anchor boxes are better queries for detr.
\newblock \emph{arXiv preprint arXiv:2201.12329}, 2022{\natexlab{a}}.

\bibitem[Liu et~al.(2021)Liu, Hu, Bai, Ding, Bai, and Torr]{liu2021multi}
Xiaolong Liu, Yao Hu, Song Bai, Fei Ding, Xiang Bai, and Philip~HS Torr.
\newblock Multi-shot temporal event localization: a benchmark.
\newblock In \emph{Proceedings of the IEEE/CVF Conference on Computer Vision and Pattern Recognition}, pages 12596--12606, 2021.

\bibitem[Liu et~al.(2022{\natexlab{b}})Liu, Wang, Hu, Tang, Zhang, Bai, and Bai]{tadtr}
Xiaolong Liu, Qimeng Wang, Yao Hu, Xu Tang, Shiwei Zhang, Song Bai, and Xiang Bai.
\newblock End-to-end temporal action detection with transformer.
\newblock \emph{IEEE Transactions on Image Processing}, 31:\penalty0 5427--5441, 2022{\natexlab{b}}.

\bibitem[Liu et~al.(2019)Liu, Ma, Zhang, Liu, and Chang]{mgg}
Yuan Liu, Lin Ma, Yifeng Zhang, Wei Liu, and Shih-Fu Chang.
\newblock Multi-granularity generator for temporal action proposal.
\newblock In \emph{Proceedings of the IEEE/CVF conference on computer vision and pattern recognition}, pages 3604--3613, 2019.

\bibitem[Loshchilov and Hutter(2017)]{adamw}
Ilya Loshchilov and Frank Hutter.
\newblock Decoupled weight decay regularization.
\newblock \emph{arXiv preprint arXiv:1711.05101}, 2017.

\bibitem[Nag et~al.(2022)Nag, Zhu, Song, and Xiang]{nag2022proposal}
Sauradip Nag, Xiatian Zhu, Yi-Zhe Song, and Tao Xiang.
\newblock Proposal-free temporal action detection via global segmentation mask learning.
\newblock In \emph{European Conference on Computer Vision}, pages 645--662. Springer, 2022.

\bibitem[Piergiovanni and Ryoo(2019)]{piergiovanni2019TGM}
AJ Piergiovanni and Michael Ryoo.
\newblock Temporal gaussian mixture layer for videos.
\newblock In \emph{International Conference on Machine learning}, pages 5152--5161. PMLR, 2019.

\bibitem[Piergiovanni and Ryoo(2018)]{piergiovanni2018superevent}
AJ Piergiovanni and Michael~S Ryoo.
\newblock Learning latent super-events to detect multiple activities in videos.
\newblock In \emph{Proceedings of the IEEE Conference on Computer Vision and Pattern Recognition}, pages 5304--5313, 2018.

\bibitem[Qing et~al.(2021)Qing, Su, Gan, Wang, Wu, Wang, Qiao, Yan, Gao, and Sang]{tcanet}
Zhiwu Qing, Haisheng Su, Weihao Gan, Dongliang Wang, Wei Wu, Xiang Wang, Yu Qiao, Junjie Yan, Changxin Gao, and Nong Sang.
\newblock Temporal context aggregation network for temporal action proposal refinement.
\newblock In \emph{Proceedings of the IEEE/CVF conference on computer vision and pattern recognition}, pages 485--494, 2021.

\bibitem[Sardari et~al.(2023)Sardari, Mustafa, Jackson, and Hilton]{sardari2023pat}
Faegheh Sardari, Armin Mustafa, Philip~JB Jackson, and Adrian Hilton.
\newblock Pat: Position-aware transformer for dense multi-label action detection.
\newblock In \emph{Proceedings of the IEEE/CVF International Conference on Computer Vision}, pages 2988--2997, 2023.

\bibitem[Shao et~al.(2023)Shao, Wang, Quan, Zheng, Yang, and Yang]{shao2023action}
Jiayi Shao, Xiaohan Wang, Ruijie Quan, Junjun Zheng, Jiang Yang, and Yi Yang.
\newblock Action sensitivity learning for temporal action localization.
\newblock \emph{arXiv preprint arXiv:2305.15701}, 2023.

\bibitem[Shi et~al.(2022)Shi, Zhong, Cao, Zhang, Ma, Li, and Tao]{react}
Dingfeng Shi, Yujie Zhong, Qiong Cao, Jing Zhang, Lin Ma, Jia Li, and Dacheng Tao.
\newblock React: Temporal action detection with relational queries.
\newblock In \emph{Computer Vision--ECCV 2022: 17th European Conference}, pages 105--121. Springer, 2022.

\bibitem[Shi et~al.(2023{\natexlab{a}})Shi, Cao, Zhong, An, Cheng, Zhu, and Tao]{shi2023temporal}
Dingfeng Shi, Qiong Cao, Yujie Zhong, Shan An, Jian Cheng, Haogang Zhu, and Dacheng Tao.
\newblock Temporal action localization with enhanced instant discriminability.
\newblock \emph{arXiv preprint arXiv:2309.05590}, 2023{\natexlab{a}}.

\bibitem[Shi et~al.(2023{\natexlab{b}})Shi, Zhong, Cao, Ma, Li, and Tao]{shi2023tridet}
Dingfeng Shi, Yujie Zhong, Qiong Cao, Lin Ma, Jia Li, and Dacheng Tao.
\newblock Tridet: Temporal action detection with relative boundary modeling.
\newblock In \emph{Proceedings of the IEEE/CVF Conference on Computer Vision and Pattern Recognition}, pages 18857--18866, 2023{\natexlab{b}}.

\bibitem[Sigurdsson et~al.(2016)Sigurdsson, Varol, Wang, Farhadi, Laptev, and Gupta]{charades}
Gunnar~A Sigurdsson, G{\"u}l Varol, Xiaolong Wang, Ali Farhadi, Ivan Laptev, and Abhinav Gupta.
\newblock Hollywood in homes: Crowdsourcing data collection for activity understanding.
\newblock In \emph{Computer Vision--ECCV 2016: 14th European Conference,}, pages 510--526. Springer, 2016.

\bibitem[Simonyan and Zisserman(2014)]{simonyan2014twostream}
Karen Simonyan and Andrew Zisserman.
\newblock Two-stream convolutional networks for action recognition in videos.
\newblock \emph{Advances in neural information processing systems}, 27, 2014.

\bibitem[Tan et~al.(2021)Tan, Tang, Wang, and Wu]{rtdnet}
Jing Tan, Jiaqi Tang, Limin Wang, and Gangshan Wu.
\newblock Relaxed transformer decoders for direct action proposal generation.
\newblock In \emph{Proceedings of the IEEE/CVF international conference on computer vision}, pages 13526--13535, 2021.

\bibitem[Tan et~al.(2022)Tan, Zhao, Shi, Kang, and Wang]{pointtad}
Jing Tan, Xiaotong Zhao, Xintian Shi, Bing Kang, and Limin Wang.
\newblock Pointtad: Multi-label temporal action detection with learnable query points.
\newblock In \emph{NeurIPS}, 2022.

\bibitem[Tan et~al.(2023)Tan, Wang, Wu, and Wang]{temporal_perceiver}
Jing Tan, Yuhong Wang, Gangshan Wu, and Limin Wang.
\newblock Temporal perceiver: A general architecture for arbitrary boundary detection.
\newblock \emph{IEEE Transactions on Pattern Analysis and Machine Intelligence}, 45\penalty0 (10):\penalty0 12506--12520, 2023.

\bibitem[Tirupattur et~al.(2021)Tirupattur, Duarte, Rawat, and Shah]{mlad}
Praveen Tirupattur, Kevin Duarte, Yogesh~S Rawat, and Mubarak Shah.
\newblock Modeling multi-label action dependencies for temporal action localization.
\newblock In \emph{Proceedings of the IEEE/CVF Conference on Computer Vision and Pattern Recognition}, pages 1460--1470, 2021.

\bibitem[Vaswani et~al.(2017)Vaswani, Shazeer, Parmar, Uszkoreit, Jones, Gomez, Kaiser, and Polosukhin]{transformer}
Ashish Vaswani, Noam Shazeer, Niki Parmar, Jakob Uszkoreit, Llion Jones, Aidan~N Gomez, {\L}ukasz Kaiser, and Illia Polosukhin.
\newblock Attention is all you need.
\newblock \emph{Advances in neural information processing systems}, 30, 2017.

\bibitem[Wang et~al.(2016)Wang, Xiong, Wang, Qiao, Lin, Tang, and Van~Gool]{tsn}
Limin Wang, Yuanjun Xiong, Zhe Wang, Yu Qiao, Dahua Lin, Xiaoou Tang, and Luc Van~Gool.
\newblock Temporal segment networks: Towards good practices for deep action recognition.
\newblock In \emph{European conference on computer vision}, pages 20--36. Springer, 2016.

\bibitem[Wang et~al.(2022)Wang, Zhang, Zheng, and Pan]{wang2022rcl}
Qiang Wang, Yanhao Zhang, Yun Zheng, and Pan Pan.
\newblock Rcl: Recurrent continuous localization for temporal action detection.
\newblock In \emph{Proceedings of the IEEE/CVF Conference on Computer Vision and Pattern Recognition}, pages 13566--13575, 2022.

\bibitem[Wang et~al.()Wang, Zhang, Yang, and Sun]{wang2109anchor}
Y Wang, X Zhang, T Yang, and J Sun.
\newblock Anchor detr: Query design for transformer-based object detection. arxiv 2021.
\newblock \emph{arXiv preprint arXiv:2109.07107}.

\bibitem[Xia et~al.(2022)Xia, Wang, Zhou, Zheng, and Tang]{xia2022learning}
Kun Xia, Le Wang, Sanping Zhou, Nanning Zheng, and Wei Tang.
\newblock Learning to refactor action and co-occurrence features for temporal action localization.
\newblock In \emph{Proceedings of the IEEE/CVF Conference on Computer Vision and Pattern Recognition}, pages 13884--13893, 2022.

\bibitem[Xu et~al.(2023{\natexlab{a}})Xu, Shen, Shi, Chen, Liao, Chen, and Wang]{xu2023progressive}
Chen Xu, Haocheng Shen, Fengyuan Shi, Boheng Chen, Yixuan Liao, Xiaoxin Chen, and Limin Wang.
\newblock Progressive visual prompt learning with contrastive feature re-formation.
\newblock \emph{arXiv preprint arXiv:2304.08386}, 2023{\natexlab{a}}.

\bibitem[Xu et~al.(2023{\natexlab{b}})Xu, Zhu, Zhang, Shen, Liao, Chen, Wu, and Wang]{xu2023dpl}
Chen Xu, Yuhan Zhu, Guozhen Zhang, Haocheng Shen, Yixuan Liao, Xiaoxin Chen, Gangshan Wu, and Limin Wang.
\newblock Dpl: Decoupled prompt learning for vision-language models.
\newblock \emph{arXiv preprint arXiv:2308.10061}, 2023{\natexlab{b}}.

\bibitem[Xu et~al.(2017)Xu, Das, and Saenko]{rc3d}
Huijuan Xu, Abir Das, and Kate Saenko.
\newblock R-c3d: Region convolutional 3d network for temporal activity detection.
\newblock In \emph{Proceedings of the IEEE international conference on computer vision}, pages 5783--5792, 2017.

\bibitem[Xu et~al.(2020)Xu, Zhao, Rojas, Thabet, and Ghanem]{g_tad}
Mengmeng Xu, Chen Zhao, David~S Rojas, Ali Thabet, and Bernard Ghanem.
\newblock G-tad: Sub-graph localization for temporal action detection.
\newblock In \emph{Proceedings of the IEEE/CVF Conference on Computer Vision and Pattern Recognition}, pages 10156--10165, 2020.

\bibitem[Xu et~al.(2019)Xu, See, and Lin]{xu2019localization}
Qichao Xu, John See, and Weiyao Lin.
\newblock Localization guided fight action detection in surveillance videos.
\newblock In \emph{2019 IEEE International Conference on Multimedia and Expo (ICME)}, pages 568--573. IEEE, 2019.

\bibitem[Yang et~al.(2019)Yang, Yang, Liu, Xiao, Davis, and Kautz]{yang2019step}
Xitong Yang, Xiaodong Yang, Ming-Yu Liu, Fanyi Xiao, Larry~S Davis, and Jan Kautz.
\newblock Step: Spatio-temporal progressive learning for video action detection.
\newblock In \emph{Proceedings of the IEEE/CVF Conference on Computer Vision and Pattern Recognition}, pages 264--272, 2019.

\bibitem[Yeung et~al.(2018)Yeung, Russakovsky, Jin, Andriluka, Mori, and Fei-Fei]{multi_thumos}
Serena Yeung, Olga Russakovsky, Ning Jin, Mykhaylo Andriluka, Greg Mori, and Li Fei-Fei.
\newblock Every moment counts: Dense detailed labeling of actions in complex videos.
\newblock \emph{International Journal of Computer Vision}, 126:\penalty0 375--389, 2018.

\bibitem[Yuan et~al.(2017)Yuan, Stroud, Lu, and Deng]{yuan2017temporal}
Zehuan Yuan, Jonathan~C Stroud, Tong Lu, and Jia Deng.
\newblock Temporal action localization by structured maximal sums.
\newblock In \emph{Proceedings of the IEEE Conference on Computer Vision and Pattern Recognition}, pages 3684--3692, 2017.

\bibitem[Zeng et~al.(2019)Zeng, Huang, Tan, Rong, Zhao, Huang, and Gan]{p_gcn}
Runhao Zeng, Wenbing Huang, Mingkui Tan, Yu Rong, Peilin Zhao, Junzhou Huang, and Chuang Gan.
\newblock Graph convolutional networks for temporal action localization.
\newblock In \emph{Proceedings of the IEEE/CVF international conference on computer vision}, pages 7094--7103, 2019.

\bibitem[Zhang et~al.(2022{\natexlab{a}})Zhang, Wu, and Li]{actionformer}
Chen-Lin Zhang, Jianxin Wu, and Yin Li.
\newblock Actionformer: Localizing moments of actions with transformers.
\newblock In \emph{Computer Vision--ECCV 2022: 17th European Conference}, pages 492--510. Springer, 2022{\natexlab{a}}.

\bibitem[Zhang et~al.(2023)Zhang, Zhu, Wang, Chen, Wu, and Wang]{zhang2023extracting}
Guozhen Zhang, Yuhan Zhu, Haonan Wang, Youxin Chen, Gangshan Wu, and Limin Wang.
\newblock Extracting motion and appearance via inter-frame attention for efficient video frame interpolation.
\newblock In \emph{Proceedings of the IEEE/CVF Conference on Computer Vision and Pattern Recognition}, pages 5682--5692, 2023.

\bibitem[Zhang et~al.(2022{\natexlab{b}})Zhang, Li, Liu, Zhang, Su, Zhu, Ni, and Shum]{dino}
Hao Zhang, Feng Li, Shilong Liu, Lei Zhang, Hang Su, Jun Zhu, Lionel Ni, and Harry Shum.
\newblock Dino: Detr with improved denoising anchor boxes for end-to-end object detection.
\newblock In \emph{International Conference on Learning Representations}, 2022{\natexlab{b}}.

\bibitem[Zhao et~al.(2021)Zhao, Thabet, and Ghanem]{vsgn}
Chen Zhao, Ali~K Thabet, and Bernard Ghanem.
\newblock Video self-stitching graph network for temporal action localization.
\newblock In \emph{Proceedings of the IEEE/CVF International Conference on Computer Vision}, pages 13658--13667, 2021.

\bibitem[Zhao et~al.(2020)Zhao, Xie, Ju, Zhang, Wang, and Tian]{zhao2020bottom}
Peisen Zhao, Lingxi Xie, Chen Ju, Ya Zhang, Yanfeng Wang, and Qi Tian.
\newblock Bottom-up temporal action localization with mutual regularization.
\newblock In \emph{Computer Vision--ECCV 2020: 16th European Conference, Glasgow, UK, August 23--28, 2020, Proceedings, Part VIII 16}, pages 539--555. Springer, 2020.

\bibitem[Zhao et~al.(2017)Zhao, Xiong, Wang, Wu, Tang, and Lin]{ssn}
Yue Zhao, Yuanjun Xiong, Limin Wang, Zhirong Wu, Xiaoou Tang, and Dahua Lin.
\newblock Temporal action detection with structured segment networks.
\newblock In \emph{Proceedings of the IEEE international conference on computer vision}, pages 2914--2923, 2017.

\bibitem[Zhao et~al.(2023)Zhao, Huang, Xing, Wu, Qiao, and Wang]{zhao2023asymmetric}
Zhiyu Zhao, Bingkun Huang, Sen Xing, Gangshan Wu, Yu Qiao, and Limin Wang.
\newblock Asymmetric masked distillation for pre-training small foundation models.
\newblock \emph{arXiv preprint arXiv:2311.03149}, 2023.

\bibitem[Zhu et~al.(2020)Zhu, Su, Lu, Li, Wang, and Dai]{deformable_detr}
Xizhou Zhu, Weijie Su, Lewei Lu, Bin Li, Xiaogang Wang, and Jifeng Dai.
\newblock Deformable detr: Deformable transformers for end-to-end object detection.
\newblock \emph{arXiv preprint arXiv:2010.04159}, 2020.

\bibitem[Zhu et~al.(2021)Zhu, Tang, Wang, Zheng, and Hua]{contextloc}
Zixin Zhu, Wei Tang, Le Wang, Nanning Zheng, and Gang Hua.
\newblock Enriching local and global contexts for temporal action localization.
\newblock In \emph{Proceedings of the IEEE/CVF international conference on computer vision}, pages 13516--13525, 2021.

\bibitem[Zhu et~al.(2022)Zhu, Wang, Tang, Liu, Zheng, and Hua]{zhu2022learning}
Zixin Zhu, Le Wang, Wei Tang, Ziyi Liu, Nanning Zheng, and Gang Hua.
\newblock Learning disentangled classification and localization representations for temporal action localization.
\newblock In \emph{Proceedings of the AAAI Conference on Artificial Intelligence}, pages 3644--3652, 2022.

\end{thebibliography}
}


\end{document}